\documentclass{article}
\usepackage{arxiv}
\usepackage[utf8]{inputenc} 
\usepackage[T1]{fontenc}    
\usepackage{hyperref}       
\usepackage{url}            
\usepackage{booktabs}       
\usepackage{amsfonts}       
\usepackage{nicefrac}       
\usepackage{microtype}      
\usepackage{lipsum}		
\usepackage{graphicx}
\usepackage{natbib}
\usepackage{doi}

\usepackage{amsmath}
\usepackage{textgreek} 
\usepackage[table]{xcolor}
\usepackage{array}
\usepackage{algorithm}
\usepackage{algpseudocode}

\title{A Hybrid CNN-LSTM Intrusion Detection Framework for Cybersecurity in Smart Renewable Energy Grids}

\author{ \href{https://orcid.org/0009-0003-2877-3136}{\includegraphics[scale=0.06]{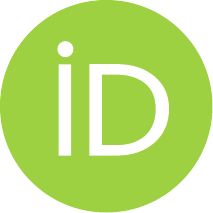}\hspace{1mm}Sajib Debnath} \\
	O\&M Analytics, AES Clean Energy\\
	The AES Corporation\\
	Louisville, CO 80027, USA \\
	\texttt{sajib.debnath@aes.com} \\
	\And
	\href{https://orcid.org/0009-0009-4052-9279}{\includegraphics[scale=0.06]{orcid.pdf}\hspace{1mm}Remon Das} \\
	Renewable Infrastructure\\
	Dominion Energy\\
	Richmond, VA 23219, USA \\
	\texttt{remondas4@gmail.com} \\
}

\hypersetup{
pdftitle={A template for the arxiv style},
pdfsubject={q-bio.NC, q-bio.QM},
pdfauthor={David S.~Hippocampus, Elias D.~Striatum},
pdfkeywords={First keyword, Second keyword, More},
}

\begin{document}
\maketitle

\begin{abstract}
The accelerated digitalization of renewable energy smart grids through IoT sensors, Advanced Metering Infrastructure (AMI), and SCADA systems has significantly expanded the attack surface for sophisticated cyberattacks, including False Data Injection (FDI) attacks that stealthily distort state estimation and DoS/DDoS attacks that flood communication channels. Current Intrusion Detection However, intrusion detection systems (IDS) exhibit three inherent limitations: inadequate modeling of the temporal progression of multi-step attacks, degraded scalability under extremely skewed class distributions of standard benchmark datasets, and restricted generalization across heterogeneous network environments. In this study, we present a Hybrid CNN-LSTM IDS that jointly exploits CNN-based spatial feature extraction and LSTM-based temporal sequence modeling, enabling the detection of instantaneous volumetric anomalies and gradually evolving low and slow-attack campaigns in real time. The model was trained using a seven-step preprocessing workflow comprising missing-value imputation, min-max normalization, one-hot encoding, SMOTE class balancing, mutual-information feature selection, causal temporal sequence construction ($T = 10$), and stratified partitioning. LSTM (96.1\%), Random Forest (93.5\%), SVM (91.2\%) and KNN (89.7\%); in NSL-KDD, it reaches 98.2\% precision versus 96.4\% (LSTM), 95.2\% (CNN), 92.7\% (Random Forest) and 90.8\% (SVM), with margins of 2--9 percentage points in all measures. An ablation analysis identified SMOTE balancing as the most influential design choice ($-3.7$~pp F1 without it). The model achieves a real-time inference throughput of 27,800~flows/s on GPU and 0.082~ms/sample CPU latency in FP32,, with INT8 quantization providing an additional $3.1\times$ speedup at 0.3\% accuracy loss, confirming deployment feasibility on resource-constrained IEDs with $<$128MB memory and establishing a deployable deep-learning framework for securing next-generation renewable energy smart grid infrastructure.
\end{abstract}

\keywords{Cybersecurity \and Smart Grid \and Deep Learning \and Hybrid CNN-LSTM \and Intrusion Detection \and False Data Injection \and DoS \and Renewable Energy \and SMOTE \and Edge Deployment}

\section{Introduction}

\label{sec:1}

We are usufructing from the largest energy transformation in history across the globe. Over the past decade, renewable energy sources (primarily solar photovoltaic (PV) and wind) have become massive shares of global electricity generation through climate policy mandates and the decreasing prices of renewable energy technologies, together with increasing awareness about security of supply. This share is expected to keep climbing rapidly throughout the next decade ~\cite{b1}, making it an increasingly large segment of our global electricity production. This extensive growth embraces smart grid infrastructure, or two-way digitally mediated energy distribution (AMI), Internet of Things (IoT) connected sensors, SCADA and Energy Management Systems (EMS). The Confluence of Deep Learning Frameworks
IoT-based sensing has significantly improved the operational efficiency of smart grids, however, attaining 93.38\% prediction accuracy for energy demand and 96.25\% accurate grid power stability~\cite{b2}, rendering potentially quite lucrative both the technology itself, and also the heavenly urgency of delivering appropriate defensive mechanisms for such AI-driven frameworks."

Smart grids have a uniquely complex risk landscape that focuses on cybersecurity. In a rational way, smart grid cyberthreats that have occurred over the years have been categorically classified into five primary modalities: Electricity Theft Attacks (ETA), False Data Injection Attacks on State Estimation (FDIA), False Command Attacks (FCA), Communication Traffic Attacks (CTA), and adversarial learning attacks (ALA) \cite{b3}. Consequently, SMG Cyber Defense has similarly been categorized into reactive methods (IDS/signature-based detection) and proactive mechanisms (IPS/MTD/DL-based anomaly detection) \cite{b4}, drawing the conceptual boundary on which the proposed model is built. Among the five threat modalities impacted by the issue of sufficient sensor observability, FDIA is particularly pernicious: an adversary equipped with partial knowledge about the power system topology can deliberately corrupt sensor measurements so that conventional bad data detection (BDD) algorithms are unable to detect them and incorrectly derived state estimates cause erroneous dispatch decisions~\cite{b5}

Research on machine learning for network intrusion detection has an established pedigree. Motivated by the statistical problems caused by duplicate record inflation in the original KDD Cup 1999 benchmark, additional work was carried out to create an updated dataset called NSL-KDD \cite{b6}, which continues to be a widely used, cross-generational benchmarking dataset. This has been reinforced through thorough comparative analyses that confirmed deep learning qualitatively outperforming traditional ML methods over a variety of network traffic contexts \cite{b7}. Intrusion detection systems based on CNNs encode features of network traffic as 1D or 2D tensors, which in turn allows convolutional filters to automatically learn discriminative local patterns and deliver a strong performance on benchmark datasets, notwithstanding their limited ability to detect slow, temporally dispersed attack patterns \cite{b8}. This temporal limit is overcome using LSTM-based approaches, which show very good detection of multi-step exploitation attacks by sequentially modeling flow behavior \cite{b9}. Hybrid CNN-LSTM architectures have been successfully studied in neighbouring domains, where they innately outperform single components.

As the most commonly adopted evaluation benchmark for modern IDS, the CICIDS2017 benchmark \cite{b10} consists of approximately 2.8 million labeled network flow records encompassing ten attack categories. It is inherently imbalanced with an extreme class imbalance across different classes, which provides a crucial limitation for fair detection across all classes in a multi-class scenario. The Synthetic Minority Oversampling Technique \cite{b11} is critical to address this imbalance to achieve consistently high recall on rare yet operationally important attack categories such as infiltration. The Adam optimizer~\cite{b12} was used for model training because it offers adaptive learning rate adjustment and has become the de-facto optimization algorithm of deep learning IDS pipelines \cite{b13}.

At the theoretical level of faithfulness, the proposed architecture is based on two landmark contributions. Specifically, deep learning, as a representational learning paradigm \cite{b14} forms the basis for automatically deriving feature hierarchies from raw network flow data with no manual effort in feature engineering. The Long Short-Term Memory (LSTM) architecture \cite{b15} overcomes the vanishing gradient issue found in traditional recurrent networks and allows the modelling of long-term temporal dependencies, a property that directly pertains to recognizing multi-step low-and-slow attack campaigns occurring through sequential smart grid traffic flows. Previous observations confirm that coordinated FDIA campaigns remain feasible against even hardened power system configurations \cite{b16}, and an extensive survey of the variety in FDIA approaches confirms that injection attacks can nearly always escape residual-based BDD given only partial topology knowledge from the attacker \cite{b17}, both of which motivate the anomaly based, data-driven detection paradigm adopted in this study.

If attacks penetrate deeper within the grid infrastructure, where they can cause physical damage to equipment, longer-duration outages, and cascading grid failures, these operational ramifications are also a national security threat, as evidenced by the 2015-2016 Ukraine power grid cyberattacks \cite{b7}, the 2021 Colonial Pipeline ransomware attack \cite{b8}, and other significant compromises of industrial safety systems, such as Triton/TRISIS health monitoring solutions \cite{b9}. Traditional ML-based IDS, such as SVM, random forest, and K-nearest neighbors, possess three major structural limitations: they do not model the time evolution of attack patterns in a dynamic environment; they require an expensive retraining process periodically; and they poorly match the real-time execution requirements needed by real systems, such as smart grids. Both standalone CNN and standalone LSTM cannot adapt to the dual spatial-temporal characteristics of smart grid network traffic~\cite{b3,b4}.

This study proposes a Hybrid CNN-LSTM Intrusion Detection System (IDS) that unifies both architectures in an end-to-end trainable model, which is rigorously validated against the CICIDS2017 and NSL-KDD benchmarks.

The main contributions of this study are summarized as follows:
\begin{itemize}
    \item \textbf{Hybrid CNN-LSTM IDS Architecture:} A novel deep learning framework that combines CNN-based spatial feature extraction with LSTM-based temporal modeling, enabling effective detection of both instantaneous and multi-stage cyberattacks.

    \item \textbf{Seven-Stage Preprocessing Pipeline:} A comprehensive data processing pipeline addressing missing values, normalization, categorical encoding, SMOTE-based class balancing, mutual information feature selection, temporal sequence construction, and stratified dataset partitioning.

    \item \textbf{Cross-Dataset Generalization:} Extensive evaluation on CICIDS2017 and NSL-KDD datasets under identical experimental settings, demonstrating strong generalization across heterogeneous network environments.

    \item \textbf{State-of-the-Art Performance:} Achieves 98.7\% accuracy and 0.995 AUC-ROC on CICIDS2017, outperforming traditional machine learning (SVM, Random Forest, KNN) and standalone deep learning (CNN, LSTM) models by 2.6--9.0 percentage points.

    \item \textbf{Ablation and Sensitivity Analysis:} Systematic evaluation of architectural and preprocessing components, identifying SMOTE-based class balancing as the most impactful factor and validating design choices.

    \item \textbf{Real-Time and Edge Deployment Feasibility:} Demonstrates high inference throughput (27,800 flows/second), low CPU latency, and INT8 quantization compatibility for deployment on resource-constrained edge devices.
\end{itemize}

\section{Related Work}
\label{sec:2}

Over time, the literature on smart grid cybersecurity has provided an organized view of the threat landscape. The most comprehensive taxonomy arranges smart grid cyberthreats into five basic categories characterized by the targets and goals of the attack, as summarized below: (1) Electricity Theft Attacks (ETAs): Individual users faking their meter readings; (2) False Data Injection Attacks on State Estimation (FDIA): Reflects corrupted state estimation and passively avoids traditional bad data detection (BDD);(3) False Command Attacks on PMU data and relay commands termed as FCA; this includes passive queries that are capable of fabricating relays’ states(e.g., boost voltage or generator powering-up modes), so as to corrupt operation decisions while at the same time bypassing routine command BDD;(4) Communication Traffic Attack appears such as Denial-of-Service attack, Distributed-Denial-of-Service attack, Time Delay Attack which is collecting enough but false observations to create wrong conclusions thus making it a symbolical formulation context communication structure built upon uncertain metrics and obscured by heavy use simulate seasonings;(5)Adversarial Learning Attack exploiting vulnerabilities within DL models that have already been deployed \cite{b3}. This taxonomy is further enriched by distinguishing between reactive (IDS, signature-based, anomaly based) and proactive cyber defense (IPS, encryption, Moving Target Defense; DL-based anomaly detection), which delineates the conceptual space in which the proposed work takes place. In addition to the unique survey on algorithms for detecting FDIA \cite{b18}, the authors further specified the mathematical bounds under which injection vectors can pass BDD and demonstrated that even detectors based purely on thresholds over residuals are inherently susceptible to topology-aware adversaries. Extending this, real-time deep-learning frameworks have shown the ability to detect tiny sensor faults in smart grid state estimation that go undetected by physics-based monitors altogether~\cite{b19}.

ETA drove research momentum in the early stages, whereas FDIA has grown exponentially since 2018 and represents such a large proportion of publications that it is now the most common topic. Consequently, ALA is an emerging category with increasing interest because of the inherent understanding that DL models leveraged for grid control are attack surfaces themselves. Among the DL mechanisms, FCL and RNN are the main streams. However, with respect to the topology characteristics of the power grid (buses = nodes, transmission lines = edges), GNN has been gaining popularity in recent years, together with AM, which is increasingly acknowledged for its ability to focus learning on the most discriminative measurement features during cyber-attack detection. These advances have strongly driven the development of principled frameworks for feature construction that derive structured representations from unstructured network flow data~\cite{b20} and established an analytical foundation for feature engineering strategies used in existing deep learning intrusion detection system (IDS) pipelines.

The KDD Cup 1999 was the first application of machine learning to network intrusion detection, resulting in many research papers investigating these classifiers (Decision Trees, Na\"{i}ve Bayes, kNN, and SVM). The KDD Cup 1999 data were discovered to have serious statistical flaws, especially regarding the inflation of reported accuracy by duplicate records, and NSL-KDD has been suggested as an improved gold-standard benchmark and is now a well-couched network intrusion detection system (IDS) evaluation dataset \cite{b5}. This traditional IDS application is based on the theoretical framework of support vector classification \cite{b21}; it has strong generalizability for linearly separable binary detection tasks but suffers from scalability constraints in high-dimensional multi-class network traffic. It was (and still is) common practice to achieve state-of-the-art performance in multi-class IDS scenarios through the use of ensemble methods such as Random Forest \cite{b22}; variance reduction from aggregated decision boundaries over multiple decorrelated trees consistently provides a detectable benefit. Recently, hybrid approaches based on rule-based algorithms and decision tree methods have similarly expanded the concept of traditional ML to specific IoT threat environments with interpretable detection logic for constrained network nodes \cite{b23}. However, all traditional ML-based IDS suffer from a structural limitation: they treat each network flow as an independent sample (thus masking the temporal ordering that is crucial for multi-stage attacks). The application of these approaches in operational smart grids is further limited by the inherent complexity of feature engineering, static decision boundaries, and high false-positive rates for rare attack categories \cite{b4}.

An extensive comparative study has established that, in general, deep learning consistently outperforms traditional ML in various network traffic scenarios \cite{b6}. CNN-based IDS represent the network traffic features in either 1D or 2D as tensors, where convolutional filters automatically learn the discriminative intervals of features. For instance, Kim et al.~\cite{b7} achieved 97.1\% accuracy on CICIDS2017 using a 1D-CNN but reported performance degradation for
slow DoS attack flows that appear individually harmless yet reveal a
periodic attack pattern when viewed as a time-series of flows. Convolutional architectures have also been shown to be effective in industrial control system (ICS) settings, since sequential sensor readings show structured spatial co-occurrence patterns that 1D filters can efficiently capture \cite{b24}. An LSTM-based IDS that mitigates the temporal constraints of network traffic data was proposed by Yin et al. They achieved 97.85\% accuracy on NSL-KDD \cite{b8} with the ability to work well against multi-step exploitation attacks. Together with GNN-based approaches that '[allow] for the graphical structure of power grids to be exploited' \cite{b1}) and '[achieve] topology-aware FDIA localization, [empowering] joint cyberattack detection and safe state recovery in SCADA systems. The RNN and LSTM families are favored to detect CTA and FDIA as they can detect correlations through time steps broken by attacks, sometimes even before any alerts are reported \cite{b4}. Statistical methods have included the use of ensemble autoencoder architectures in an approach to online network intrusion detection, where normal traffic distributions are learned in an unsupervised manner, and statistical deviations are detected as potential intrusions without dependence on labelled attack data \cite{b36}. Complementary anomaly detection approaches that explicitly integrate energy domain defense data with AI models have also been proposed for renewable energy systems \cite{b26}. In addition, transformer-based IDS were reviewed that emphasized the attention mechanism to learn long-range temporal dependency more efficiently than LSTMs in some anomaly detection scenarios, although they had a considerable computational cost \cite{b4,b27,b37, b39}. Recent Bayesian transformer architectures have further extended this direction by integrating epistemic uncertainty estimation directly into the attention mechanism for grid time-series forecasting, suggesting a natural path toward uncertainty-aware IDS. Including hybrid CNN-LSTM architectures in neighboring domains, which have been shown to provide better results than independent components, Wang et al. CNN-LSTM models (via ~\cite{b9}) were proposed on NSL-KDD yielding 98.1\% accuracy but without both CICIDS2017 or the data preprocessing pipeline components, especially SMOTE class balancing, which is shown to be critical for reliable high-performance levels across rare attack categories.

An essential conceptual difference between the reactive and proactive approaches in the cyber defense of smart grids is that reactive methods detect and respond to a threat after it has occurred, whereas proactive defense anticipates and prevents threats before they occur \cite{b4}. The approach adopted in this study for DL-based anomaly detection is both proactive and reactive; since it learns the normal traffic distributions and detects statistical deviations in real time, it allows early stage detection and prevention rather than post-hoc forensic analysis, thus marking a difference with other methodologies. Policy-driven adaptive mechanisms significantly boost detection robustness against FDIA injection campaigns dynamically crafted over the system trajectory, a feature revealed using attention-aware deep reinforcement learning \cite{b38}. The Moving Target Defense (MTD) approach is a proactive defense strategy that randomizes system configurations to disorient attackers and has been mathematically established as a strong optimization problem for protecting the power grid from coordinated FDIA \cite{b28,b29}. Additionally, cooperative inference has been shown to improve the detection success rate of low-volume, distributed attack patterns, avoiding centralized monitors \cite{b30} by using a demand-side defense model based on the deployment of deep learning-based distributed attack detector schemes on IoT network nodes.

Table \ref{tab:1} presents a structured comparison of representative IDS studies
from the literature, synthesizing findings from the reference
surveys~\cite{b2,b3,b4} alongside landmark primary
works~\cite{b5,b6,b7,b8,b9,b13}. This synthesis directly motivates the
design decisions for the proposed framework.

\begin{table*}[htbp]
\centering
\caption{Comparative summary of representative IDS approaches.}
\label{tab:1}
\renewcommand{\arraystretch}{1.3}
\resizebox{\textwidth}{!}{
\begin{tabular}{|p{2.5cm}|p{1.0cm}|p{2.5cm}|p{3.5cm}|p{1.0cm}|p{5cm}|}
\hline

\rowcolor{gray!15}
\textbf{Author(s)} &
\textbf{Year} &
\textbf{Dataset} &
\textbf{Method} &
\textbf{Acc.(\%)} &
\textbf{Key Limitation} \\
\hline

Liu et al.~\cite{b13} & 2011 & Synthetic &
FDI Mathematical Analysis & N/A &
Cannot detect coordinated FDI bypassing BDD \\
\hline

Tavallaee et al.~\cite{b5} & 2009 & NSL-KDD &
SVM, Decision Tree & $\sim$91\% &
No temporal modeling; static feature sets \\
\hline

Yin et al.~\cite{b8} & 2017 & NSL-KDD &
LSTM (RNN) & 97.85\% &
LSTM only; no spatial feature extraction \\
\hline

Kim et al.~\cite{b7} & 2017 & CICIDS2017 &
1D-CNN & 97.1\% &
Poor slow-DoS detection; no temporal context \\
\hline

Vinayakumar et al.~\cite{b6} & 2019 & NSL-KDD/UNSW &
MLP, CNN, LSTM (separate) & $\sim$97\% &
Models tested independently; no hybrid fusion \\
\hline

Wang et al.~\cite{b9} & 2018 & NSL-KDD &
CNN-LSTM (HAST-IDS) & 98.1\% &
NSL-KDD only; no SMOTE; no smart grid focus \\
\hline

Singh et al.~\cite{b2} & 2025 & IoT simulation &
ORA-DL (DNN+RL+MAS) & 93.4\% &
Energy optimization; not IDS-specific \\
\hline

Ruan et al.~\cite{b3} & 2023 & Multi-dataset &
Survey: FDIA/ALA/CTA & Varies &
Review only; no unified hybrid model proposed \\
\hline

Abdi et al.~\cite{b4} & 2024 & Multi-dataset &
Survey: DL proactive & Varies &
Survey only; DL-MTD integration nascent \\
\hline

\rowcolor{green!15}
\textbf{Proposed (Ours)} & \textbf{2026} &
\textbf{CICIDS2017+NSL-KDD} &
\textbf{Hybrid CNN-LSTM+SMOTE} & \textbf{98.7\%} &
\textbf{Full multi-class; cross-dataset; domain-specific} \\
\hline

\end{tabular}
}
\vspace{2mm}
\end{table*}

Three key research gaps motivate the present work: (1) absence of smart
grid domain-specific adaptation in existing hybrid CNN-LSTM IDS; (2)
incomplete preprocessing particularly neglect of severe class imbalance
in CICIDS2017~\cite{b10} that causes high overall accuracy but poor
recall on rare critical attack types, a limitation that SMOTE~\cite{b11}
directly addresses; and (3) limited cross-dataset validation across
heterogeneous network environments. Multi-dimensional feature fusion
combined with stacking ensemble mechanisms has been shown to improve
detection robustness under imbalanced and high-dimensional
settings~\cite{b31}, reinforcing the value of systematic feature
selection in the proposed pipeline. The breadth of available IDS
evaluation benchmarks spanning KDD Cup, NSL-KDD, CICIDS2017, and IoT
datasets were systematically reviewed ~\cite{b32}, highlighting the absence of SCADA-native labeled datasets as a persistent limitation requiring synthetic traffic generation. Machine learning-based vulnerability analysis of Industrial Internet of Things (IIoT) deployments~\cite{b33} has demonstrated that network-layer behavioral features transfer across IoT and operational technology (OT)
environments, supporting the cross-domain applicability of the proposed
model. Federated learning frameworks for intrusion detection in industrial cyber-physical systems~\cite{b34} represent a complementary direction, enabling privacy-preserving collaborative training across
distributed grid nodes without centralizing the raw traffic data. The
security and privacy trade-offs of federated IDS have been examined
comprehensively~\cite{b35}, identifying gradient leakage and model
poisoning attacks as open challenges for federated smart grid deployment. This paper addresses the three identified gaps through a domain-informed, The SMOTE-enhanced, cross-validated Hybrid CNN-LSTM IDS was explicitly designed for renewable energy smart grid security.

\section{System Model and Threat Environment}
\label{sec:3}

\begin{figure}[t]	
\centering
\includegraphics[width=1.0\textwidth]{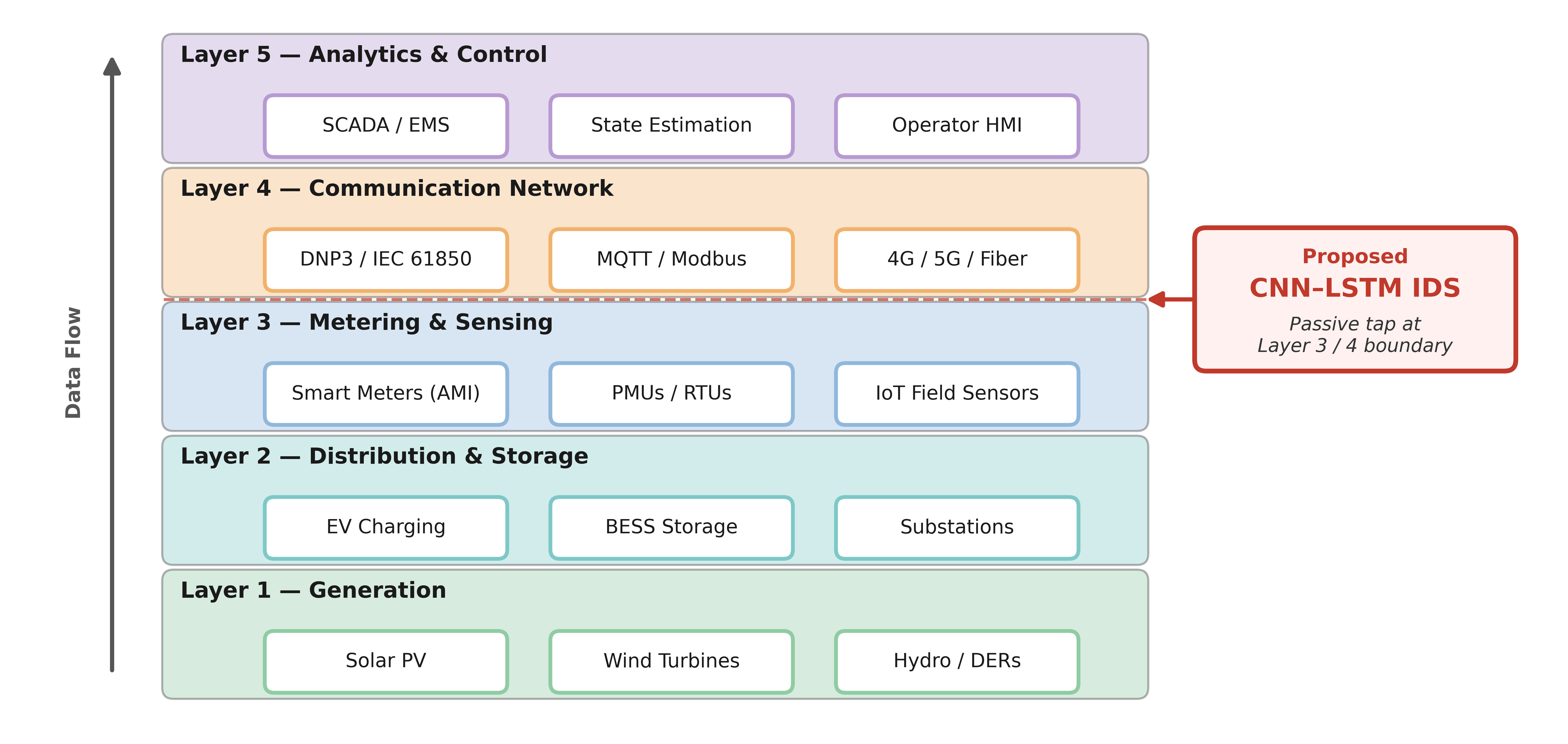}
\caption{Five-layer smart renewable energy grid architecture with the
proposed CNN-LSTM IDS deployed as a passive tap at the Layer~3/4 boundary.}
\label{fig:1}
\end{figure}

Figure~\ref{fig:1} illustrates the proposed system model: a five-layer hierarchical smart renewable energy grid, consistent with the OT/IT convergence architecture of Singh et al. ~\cite{b2} and layered SG infrastructure taxonomy of Abdi et al. ~\cite{b4}. Layer 1 covers renewable generation (solar PV, wind, hydro/DERs); Layer 2 covers distribution and storage (EV charging, BESS, substations); Layer 3 is the metering and sensing layer (AMI meters, PMUs/RTUs, IoT sensors) and serves as the primary IDS data collection point; Layer 4 is the heterogeneous communication network (DNP3/IEC 61850, MQTT/Modbus, 4G/5G/fiber); and Layer 5 is the analytics and control domain (SCADA/EMS, state estimation, operator HMI). The proposed CNN-LSTM IDS is deployed as a passive tap at the Layer 3/4 boundary ~\cite{b4}, monitoring the measurement traffic entering the communication network without introducing operational latency.

The proposed CNN-LSTM IDS is deployed at the boundary between Layers 3 and 4, where all field measurement data enter the communication network bound for the control center. Network taps or port mirroring on Layer 3 switches provide the IDS with a passive copy of all traffic without introducing latency into the operational data path. This placement is consistent with Network-based IDS (NIDS) architectures, as classified by Abdi et al.~\cite{b4}, which monitor network traffic to identify intrusions targeting nodes or devices without interrupting operational communication.

We adopt a threat model consistent with the advanced persistent threat (APT) paradigm relevant to nation-state-level attacks on critical infrastructure, covering all five attack categories identified by Ruan et al.~\cite{b3}. The adversary is assumed to possess the following characteristics: (1) network access to the smart grid communication infrastructure; (2) partial knowledge of the power system topology
sufficient to construct FDI attacks bypassing the BDD; (3) control of a botnet capable of generating DDoS traffic exceeding the target communication bandwidth; and (4) capability to execute multi-stage, low-and-slow campaigns.

The power system state estimation model underlying characteristic~(2) is expressed as:
\begin{equation}
    \mathbf{z} = \mathbf{H}\mathbf{x} + \mathbf{e}
    \label{eq:se}
\end{equation}
where $\mathbf{z} \in \mathbb{R}^{m}$ is the measurement vector,
$\mathbf{H} \in \mathbb{R}^{m \times n}$ is the known system matrix,
$\mathbf{x} \in \mathbb{R}^{n}$ is the true state vector, and
$\mathbf{e} \sim \mathcal{N}(\mathbf{0}, \mathbf{R})$ is Gaussian measurement noise.
Traditional Bad Data Detection (BDD) raises an alarm when the weighted residual
exceeds a threshold $\tau$:
\begin{equation}
    \|\mathbf{z} - \mathbf{H}\hat{\mathbf{x}}\|_{2} > \tau
    \label{eq:bdd}
\end{equation}
An FDI attack vector $\mathbf{a}$ is \emph{stealthy} if and only if
$\mathbf{a} = \mathbf{H}\mathbf{c}$ for some nonzero perturbation
$\mathbf{c} \in \mathbb{R}^{n}$, since the corrupted residual satisfies
$\|(\mathbf{z}+\mathbf{a}) - \mathbf{H}\hat{\mathbf{x}}_{a}\|_{2}
= \|\mathbf{z} - \mathbf{H}\hat{\mathbf{x}}\|_{2} \leq \tau$,
bypassing BDD entirely precisely the scenario motivating the data-driven
detection approach proposed in this study.

\begin{table*}[htbp]
\centering
\caption{Threat model aligned with the five-category smart grid
cyberthreat taxonomy of Ruan et al.~\cite{b3}.}
\label{tab:2}

\renewcommand{\arraystretch}{1.2}

\resizebox{\textwidth}{!}{
\begin{tabular}{|p{2.5cm}|p{4.5cm}|p{4cm}|p{3cm}|}
\hline

\rowcolor{gray!15}
\textbf{Threat Category} &
\textbf{Description} &
\textbf{Grid Target} &
\textbf{Modality per Ruan et al. \cite{b3}} \\
\hline

FDI Attack & Systematic sensor data manipulation to bypass BDD & State Estimator (Layer 5) & FDIA \\
\hline

DoS / DDoS & Flooding communication channels with spurious traffic & Communication (Layer 4) & CTA \\
\hline

Brute Force & Repeated authentication attempts on SCADA interfaces & Control Center (Layer 5) & FCA \\
\hline

Botnet / Malware & Coordinated IoT device compromise and C\&C activity & All Layers (IoT nodes) & CTA / FCA \\
\hline

Electricity Theft & Meter reading manipulation by prosumers & Metering Layer (Layer 3) & ETA \\
\hline
\end{tabular}
}
\vspace{2mm}
\end{table*}

Table \ref{tab:2} formalizes the threat model by mapping each concrete attack vector considered in this study to the five-category smart-grid cyber threat taxonomy of Ruan et al. \cite{b3}. Five representative threat classes, FDI, DoS/DDoS, brute force, botnet/malware, and electricity theft, are enumerated, together with their primary grid target layer and corresponding taxonomy modality (FDIA, CTA, FCA, ETA). This explicit alignment ensures that the proposed IDS spans all five recognized threat modalities across the metering (Layer 3), communication (Layer 4), and control-center (Layer 5) tiers of the smart renewable energy grid architecture depicted in Figure \ref{fig:1}.

\section{Proposed Methodology}
\label{sec:4}

\subsection{Framework Overview} 

\begin{figure}[htbp]
    \centering
    \includegraphics[width=1.0\textwidth]{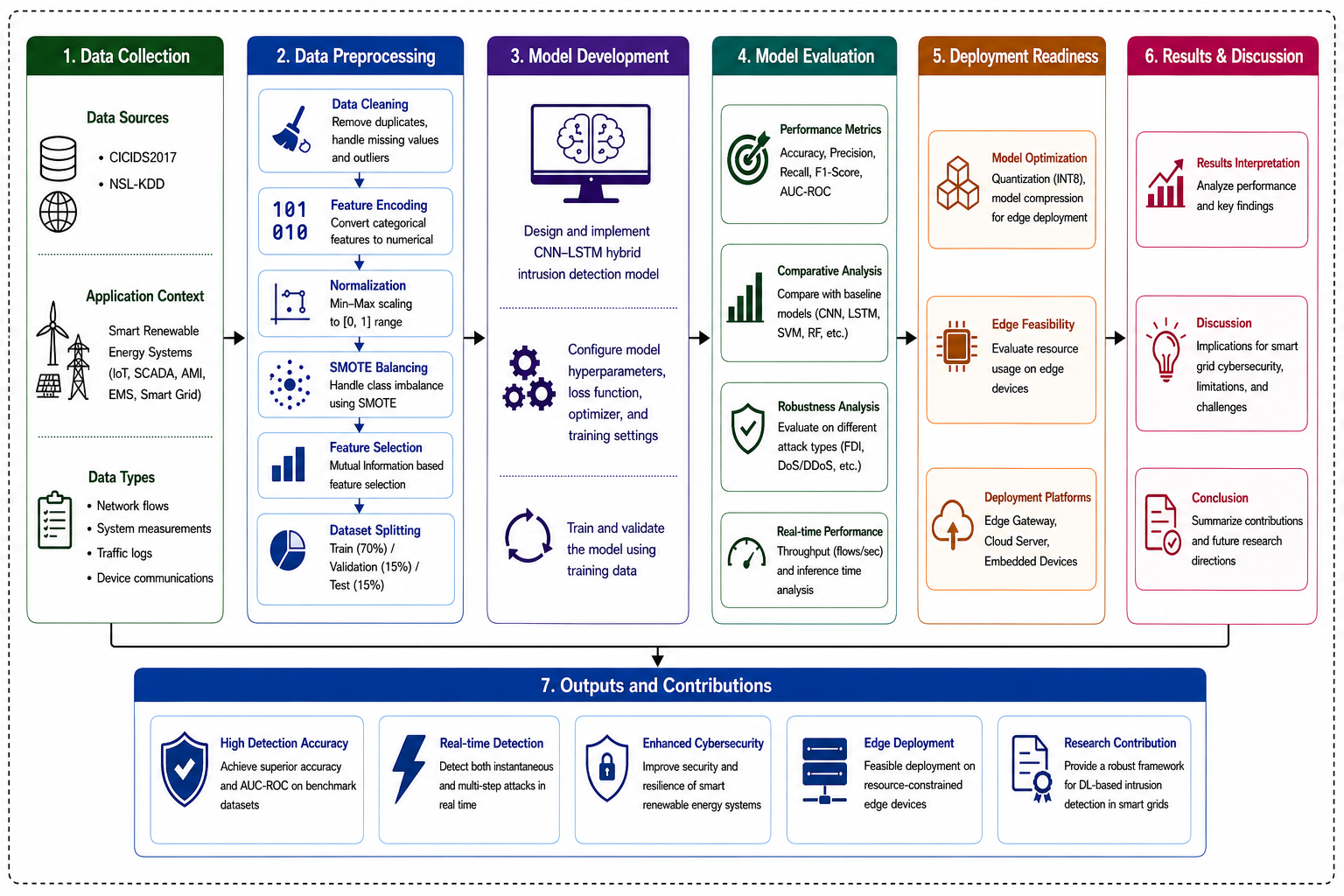}
    \caption{End-to-end workflow of the proposed CNN-LSTM IDS framework
from data collection through preprocessing, training, and edge deployment.}
    \label{fig:2}
\end{figure}

The proposed Hybrid CNN-LSTM IDS framework is structured as a seven-stage 
end-to-end pipeline, as shown in Figure~\ref{fig:2}. Each stage 
is designed with explicit consideration of the computational constraints, 
real-time latency requirements, and class imbalance challenges inherent to 
smart renewable energy grid environment.

The pipeline begins with data collection from CICIDS2017 and NSL-KDD, followed by a
seven-step preprocessing sequence data cleaning, stratified partitioning, min-max normalization, categorical encoding, SMOTE class balancing, mutual information feature selection retaining the top
40 features, and temporal sequence construction with normalization, SMOTE, and feature selection steps fitted exclusively on the training fold to prevent data leakage. The preprocessed data train the Hybrid CNN-LSTM model using the Adam optimizer at $\eta{=}0.001$
with early stopping and ReduceLROnPlateau scheduler. Model evaluation covers standard performance metrics, comparative analysis against five baselines, per-class robustness analysis, and real-time
throughput characterization. Finally, deployment readiness is confirmed via INT8 quantization ($3.1\times$ speedup, 0.3\% accuracy loss) and edge feasibility analysis on ARM-based IEDs with $<$128MB.

\subsection{Datasets}
\label{sec:5}
\subsubsection{CICIDS2017}
\label{sec:5.1}
The CICIDS2017 dataset [10] contains approximately 2.8 million labeled network flow records generated over five days using the CICFlowMeter tool, characterized by 78 features per record. Attack categories span DoS Hulk, DDoS, DoS GoldenEye, DoS Slowloris, FTP/SSH Brute Force, Web Attacks (XSS, SQLi), Botnet ARES, and Infiltration, alongside benign traffic. A critical challenge is the extreme class imbalance: benign traffic constitutes over 80\% of records, whereas infiltration accounts for fewer than 0.01\%. This imbalance was addressed by SMOTE in the preprocessing pipeline. CICIDS2017 is the most widely used modern IDS benchmark, referenced in both Ruan et al. \cite{b3} and Abdi et al. \cite{b4} as the primary standard for contemporary IDS evaluation.

\subsubsection{NSL-KDD}
\label{sec:5.2}
NSL-KDD \cite{b5} was used for cross-dataset generalization evaluation. It contains 125,973 training and 22,544 test records with 41 features organized into DoS, Probe, R2L, and U2R attack meta-classes. NSL-KDD eliminates duplicate records from the KDD Cup 1999, providing more realistic class distributions. Abdi et al. \cite{b4} noted that NSL-KDD remains valuable despite its age because it serves as a standard cross-generational benchmark, enabling comparisons with the large body of pre-2018 IDS literature.

\subsection{Seven-Stage Preprocessing Pipeline}
\label{sec:5.3}

\begin{itemize}

\item \textbf{Missing and Infinite Value Handling:}
Features with $>$5\% NaN values are dropped; remaining NaN values are
mean-imputed. Infinite values are replaced with the column maximum prior
to normalization.

\item \textbf{Min-Max Normalization:}
All numerical features are scaled to $[0,1]$ via
\begin{equation}
    x_{\text{norm}} = \frac{x - x_{\min}}{x_{\max} - x_{\min}}
    \label{eq:minmax}
\end{equation}
computed on the training split only and applied identically to the
validation and test splits.

\item \textbf{Categorical Encoding:}
Protocol type and TCP flag features are one-hot encoded to preserve
non-ordinal categorical structure.

\item \textbf{Class Balancing via SMOTE~\cite{b11}:}
SMOTE is applied exclusively to the training split. For each minority-class
sample $\mathbf{x}_i$, a synthetic neighbour is generated as:
\begin{equation}
    \mathbf{x}_{\text{syn}} = \mathbf{x}_i
    + \lambda\,\bigl(\tilde{\mathbf{x}}_i - \mathbf{x}_i\bigr),
    \quad \lambda \sim \mathcal{U}(0,1),
    \label{eq:smote}
\end{equation}
where $\tilde{\mathbf{x}}_i$ is a randomly selected $k$-nearest neighbour
of $\mathbf{x}_i$ in feature space, oversampling minority attack categories
to a balanced class ratio of $1.0$ relative to the majority class.

\item \textbf{Mutual Information Feature Selection:}
The mutual information between each feature $X_j$ and the class label $Y$
is computed as:
\begin{equation}
    I(X_j;\,Y) = \sum_{x \in \mathcal{X}}\sum_{y \in \mathcal{Y}}
    p(x,y)\,\log\frac{p(x,y)}{p(x)\,p(y)}
    \label{eq:mi}
\end{equation}
and the top $k{=}40$ features ranked by $I(X_j;Y)$ are retained,
eliminating redundant and low-discriminative dimensions from both datasets.

\item \textbf{Temporal Sequence Construction:}
After the CNN independently processes each flow to produce a
128-dimensional embedding $\mathbf{e}_t \in \mathbb{R}^{128}$,
a causal temporal sequence of $T{=}10$ consecutive embeddings is assembled as:
\begin{equation}
    \mathbf{X}_t^{\text{seq}} =
    \bigl[\mathbf{e}_{t-T+1},\;\mathbf{e}_{t-T+2},\;\ldots,\;\mathbf{e}_t\bigr]
    \in \mathbb{R}^{T \times 128}
    \label{eq:seq}
\end{equation}
which forms the input to the LSTM layer at inference step $t$.
Each flow's 40-feature vector is first reshaped to $(40, 1)$ for
1D-CNN processing.

\item \textbf{Stratified Train/Validation/Test Split:}
A 70\%/15\%/15\% stratified split by class label is performed
\emph{first}, before any fitting step, to prevent data leakage; the
test set retains representation of all attack categories.

\end{itemize}

\subsection{Hybrid CNN-LSTM Architecture}

The proposed model integrates 1D-CNN spatial feature extraction with LSTM
temporal sequence modeling. This design directly addresses the recommendation
by Ruan et al.~\cite{b3} that effective FDIA and CTA detection requires
combining RNN-family temporal modeling with CNN-based local anomaly extraction.

The CNN component learns local co-occurrence patterns across adjacent network
flow features. The output of the $k$-th filter in the first convolutional layer
is:
\begin{equation}
    y_k[t] = \sigma\!\left(\sum_{m=0}^{K-1} W_k[m]\cdot x[t+m] + b_k\right)
    \label{eq:conv}
\end{equation}
where $x[t]$ is the input feature at position $t$, $W_k \in \mathbb{R}^{K}$
is the learnable kernel of size $K{=}3$, $b_k$ is the bias, and
$\sigma(\cdot)$ denotes the ReLU activation.

The LSTM then reads the temporal sequence of CNN-derived embeddings
$\{\mathbf{e}_1, \mathbf{e}_2, \ldots, \mathbf{e}_T\}$ (\
Equation~\eqref{eq:seq}), tracking how network behavior evolves across
$T{=}10$ consecutive flows through gated state updates:
\begin{align}
    \mathbf{f}_t &= \sigma\!\bigl(\mathbf{W}_f[\mathbf{h}_{t-1}
        \mathbf{e}_t] + \mathbf{b}_f\bigr) \label{eq:fg}\\[3pt]
    \mathbf{i}_t &= \sigma\!\bigl(\mathbf{W}_i[\mathbf{h}_{t-1}
        \mathbf{e}_t] + \mathbf{b}_i\bigr) \label{eq:ig}\\[3pt]
    \tilde{\mathbf{C}}_t &= \tanh\!\bigl(\mathbf{W}_C[\mathbf{h}_{t-1}
        \mathbf{e}_t] + \mathbf{b}_C\bigr) \label{eq:cand}\\[3pt]
    \mathbf{C}_t &= \mathbf{f}_t \odot \mathbf{C}_{t-1}
        + \mathbf{i}_t \odot \tilde{\mathbf{C}}_t \label{eq:cell}\\[3pt]
    \mathbf{o}_t &= \sigma\!\bigl(\mathbf{W}_o[\mathbf{h}_{t-1}
        \mathbf{e}_t] + \mathbf{b}_o\bigr) \label{eq:og}\\[3pt]
    \mathbf{h}_t &= \mathbf{o}_t \odot \tanh(\mathbf{C}_t) \label{eq:hid}
\end{align}
where $\mathbf{f}_t$, $\mathbf{i}_t$, $\mathbf{o}_t$ are the forget, input,
and output gates; $\mathbf{C}_t$ is the cell state; and $\odot$ denotes
the Hadamard product. The forget gate in Equation~\eqref{eq:fg} directly
enables long-range dependency modeling, addressing the vanishing gradient
limitation of vanilla RNNs~\cite{b15}.

Given the final LSTM hidden state $\mathbf{h}_T \in \mathbb{R}^{128}$,
the multi-class output probability over $C$ attack categories is computed via
the softmax function:
\begin{equation}
    P\!\left(\hat{y} = k \mid \mathbf{X}^{\text{seq}}\right) =
    \frac{\exp(z_k)}{\displaystyle\sum_{j=1}^{C}\exp(z_j)}
    \quad k = 1,\ldots,C
    \label{eq:softmax}
\end{equation}
where $\mathbf{z} = \mathbf{W}_{\text{out}}\mathbf{h}_T + \mathbf{b}_{\text{out}}$
are the pre-softmax logits produced by the final dense layer.

\begin{figure}[htbp]
    \centering
    \includegraphics[width=1.0\textwidth]{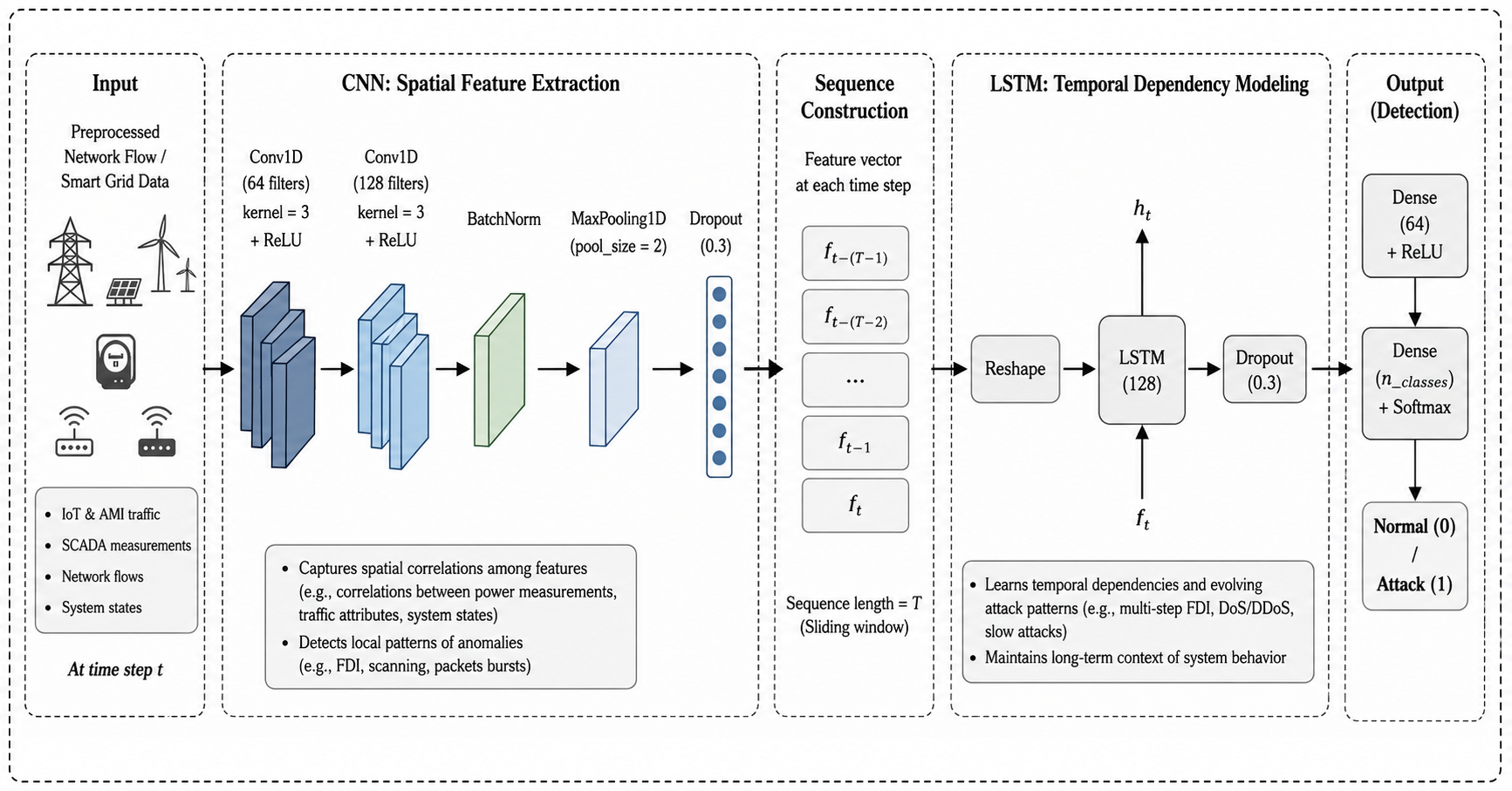}
    \caption{Proposed Hybrid CNN-LSTM IDS architecture with dual
    1D-convolutional blocks, a reshape bridge, a 128-unit LSTM, and a
    multi-class softmax output.}
    \label{fig:3}
\end{figure}

Table~\ref{tab:3} presents the complete layer-by-layer specifications of the proposed architecture. The network comprises two 1D-Conv blocks (64 and 128 filters, kernel size~3) with batch normalization, max-pooling, and dropout regularization, followed by GlobalMaxPooling1D bridge that produces a
128-dimensional per-flow embedding. Ten such embeddings (one per time step) were stacked externally to form the LSTM input sequence of shape (batch, 10, 128), which was then processed by a 128-unit LSTM and two dense layers culminating in a softmax classifier. The model totals approximately 166K trainable parameters, with measured inference latencies of 2.3~ms/batch on GPU and 8.2~ms/batch (0.180~ms/sample at batch size~100) on CPU, confirming suitability for real-time
deployment.

\begin{table*}[htbp]
\centering
\caption{Layer-by-layer specification of the proposed Hybrid CNN-LSTM IDS
($\sim$166K parameters; GPU: 2.3~ms/batch at batch~64;
  CPU: 8.2~ms/batch = 0.082~ms/sample at batch~100).}
\label{tab:3}
\renewcommand{\arraystretch}{1.2}
\resizebox{\textwidth}{!}{
\begin{tabular}{|p{1cm}|p{3.9cm}|p{2.3cm}|p{1.7cm}|p{3.5cm}|}
\hline
\rowcolor{gray!15}
\textbf{Layer} &
\textbf{Type} &
\textbf{Output Shape} &
\textbf{Parameters} &
\textbf{Notes} \\
\hline
1 & Input (features, 1 channel) & (batch, 40, 1) & 0 & Input size = 40 \\
\hline
2 & Conv1D (64 filters, ReLU) & (batch, 38, 64) & 256 & Kernel = 3 \\
\hline
3 & Batch Normalization & (batch, 38, 64) & 256 & Stabilizes activations \\
\hline
4 & MaxPooling1D & (batch, 19, 64) & 0 & Pool size = 2 \\
\hline
5 & Conv1D (128 filters, ReLU) & (batch, 17, 128) & 24,704 & Kernel = 3 \\
\hline
6 & Batch Normalization & (batch, 17, 128) & 512 & Stabilizes activations \\
\hline
7 & Dropout (0.3) & (batch, 17, 128) & 0 & Regularization \\
\hline
8 & GlobalMaxPooling1D & (batch, 128) & 0 & CNN→LSTM bridge; produces per-flow embedding \\
\hline
9 & LSTM (128 units) [input: stacked T=10 embeddings] & (batch, 128) & 131,584 & Final sequence encoding \\
\hline
10 & Dropout (0.3) & (batch, 128) & 0 & Regularization \\
\hline
11 & Dense (64, ReLU) & (batch, 64) & 8,256 & Intermediate layer \\
\hline
12 & Output (Softmax) & (batch, $n_{\text{classes}}$) & varies & Multi-class classification \\
\hline
\end{tabular}
}
\end{table*}

\subsection{Algorithm: CNN-LSTM IDS Training and Inference}
In Algorithm~\ref{alg:1}, we provide the pseudocode which schematically represents the whole CNN-LSTM IDS pipeline, including preprocessing, model training and real-time inference. The design is heavily influenced by the ORA-DL pseudocode paradigm by (Singh et al., 2018) [2] a research article that is modified for the context of IDS intrusion detection. The four phases of the algorithm are output in order: (1) preprocessing (lines 2–9): applies the seven-stage pipeline detailed in Section 4.3; (2) model construction (lines 11–15): instantiates CNN-LSTM layers; (3) training (lines 17–31): conducts Adam optimization with early stopping and learning-rate scheduling; and, lastly, real-time inference (lines 33–45), where a sliding T=10 buffer tracks if non-benign is recognized as the most probable class so alerts will trigger.

\begin{algorithm}[htbp]
\caption{Hybrid CNN-LSTM IDS -- Training and Real-Time Inference}
\label{alg:1}

\textbf{Input:} Raw network flow records $D = \{(x_i, y_i)\}$ from CICIDS2017 / NSL-KDD \\
\textbf{Output:} Trained CNN-LSTM model $M$; real-time predictions $\hat{y}$

\begin{algorithmic}[1]

  \State \textbf{Phase 1: Preprocessing (split-first; no leakage)}
  \State $(D_{\text{train}}, D_{\text{val}}, D_{\text{test}}) \gets$ StratifiedSplit($D$, 0.70, 0.15, 0.15) \Comment{Split FIRST}
  \State $D_{\text{train}} \gets$ RemoveNaNInfinite($D_{\text{train}}$); apply same mask to $D_{\text{val}}, D_{\text{test}}$ \Comment{Fit on train only}
  \State $D \gets$ MinMaxNormalize($D$, fit = TRAIN\_ONLY) \Comment{Scaler fit on $D_{\text{train}}$ only}
  \State $D \gets$ OneHotEncode($D$, cols=[protocol, flags])
  \State $D_{\text{train}} \gets$ SMOTE($D_{\text{train}}$, ratio=1.0) \Comment{Training split only; no leakage}
  \State $features \gets$ MutualInfoTopK($D_{\text{train}}, k=40$) \Comment{Fit on train only}
  \State $D_{\text{train}}, D_{\text{val}}, D_{\text{test}} \gets D[features]$
  \State $D_{\text{train}}, D_{\text{val}}, D_{\text{test}} \gets$ ReshapeSequences($T{=}10$)

\vspace{2mm}
\State \textbf{Phase 2: Model Construction}
\State $M \gets$ BuildModel():
\State \hspace{5mm} Conv1D(64, k=3, ReLU), BatchNorm, MaxPool(2)
\State \hspace{5mm} Conv1D(128, k=3, ReLU), BatchNorm, Dropout(0.3)
\State \hspace{5mm} Reshape(T=10), LSTM(128), Dropout(0.3)
\State \hspace{5mm} Dense(64, ReLU), Dense($n_{\text{classes}}$, Softmax)

\vspace{2mm}
\State \textbf{Phase 3: Training}
\State optimizer $\gets$ Adam(lr=0.001)
\State loss $\gets$ CategoricalCrossEntropy()
\State scheduler $\gets$ ReduceLROnPlateau()
\For{epoch = 1 to max epochs}
    \For{batch $B$ in $D_{\text{train}}$}
        \State $\hat{y} \gets M(B)$
        \State $L \gets$ loss($\hat{y}$, $y$)
        \State Backpropagate and update parameters
    \EndFor
    \State val\_loss $\gets$ Evaluate($M$, $D_{\text{val}}$)
    \State scheduler.step(val\_loss)
    \If{EarlyStopping(val\_loss)}
        \State \textbf{break}
    \EndIf
\EndFor

\vspace{2mm}
\State \textbf{Phase 4: Real-Time Inference}
\State Initialize buffer of size $T=10$
\While{new flow $f$ arrives} \Comment{Apply saved scaler/mask}
    \State buffer.append($f$) \Comment{ Maintain T=10 window}
    \If{len(buffer) == $T$}
        \State $x_{\text{seq}} \gets$ ReshapeSequence(buffer) \Comment{Shape: (1, 40, 1) x T}
        \State $\hat{y} \gets M(x_{\text{seq}})$ \Comment{Softmax probabilities}
        \State alert\_class $\gets \arg\max(\hat{y})$ \Comment{Highest-prob class}
        \If{alert\_class $\neq$ BENIGN}
            \State TriggerAlert(alert\_class, confidence)
        \EndIf
        \State buffer.pop(0) \Comment{slide window by 1}
    \EndIf
\EndWhile

\end{algorithmic}
\end{algorithm}

\subsection{Training Configuration}

The model is optimized by minimizing the categorical cross-entropy loss:
\begin{equation}
    \mathcal{L} = -\frac{1}{N}\sum_{i=1}^{N}\sum_{k=1}^{C}
    y_{ik}\,\log\hat{y}_{ik}
    \label{eq:loss}
\end{equation}
where $N$ is the number of training samples, $C$ is the number of classes,
$y_{ik} \in \{0,1\}$ is the one-hot ground-truth label, and $\hat{y}_{ik}$
is the predicted softmax probability for class $k$ from
Equation~\eqref{eq:softmax}. The Adam optimizer~\cite{b12} is used with
lr$= 0.001$, $\beta_1{=}0.9$, $\beta_2{=}0.999$.
A ReduceLROnPlateau scheduler reduces the learning rate by $0.5\times$ when
the validation loss stagnates for three consecutive epochs; early stopping
with patience$= 5$ prevents overfitting. Maximum training was capped at 100
epochs (typical convergence: epochs 30--50). The batch size was 64, and the
70/15/15 split was stratified by class label. These hyperparameters were
validated through a sensitivity analysis by varying lr ($0.0001$--$0.01$) and
batch size ($16$--$64$), confirming the selected values as optimal for
training stability and prediction accuracy, consistent with the empirical
validation approach of Singh et al.~\cite{b2}.

\section{Experimental Results}
\label{sec:5}

\subsection{Experimental Setup}
\label{sec:5.1}
All experiments are implemented in Python~3.9 using TensorFlow~2.12/Keras, scikit-learn~1.3, and imbalanced-learn for SMOTE. Hardware: NVIDIA RTX~3080 GPU (10~GB VRAM) and Intel Core i7-12700K CPU. The evaluation protocol is 5-fold stratified cross-validation: the full dataset is partitioned into five stratified folds; within each fold the preprocessing pipeline (normalization,
SMOTE, and MI feature selection) is fitted exclusively on the training portion and applied to the held-out
test fold. A 15\% validation split within each fold was used solely for early stopping and learning rate scheduling. All reported metrics are averages of the five test folds to eliminate partition-dependent variance.
Baseline hyperparameters are selected via grid search: SVM (RBF kernel, $C{=}10$, $\gamma{=}0.01$), Random Forest ($n_{\text{estimators}}{=}200$, $\text{max\_depth}{=}20$), and KNN ($k{=}5$). The standalone CNN and LSTM baselines used the identical seven-stage preprocessing pipeline as the proposed model, differing only in architecture, ensuring a controlled architectural comparison.
Model performance is quantified per class $k$ using:
\begin{align}
    \text{Precision}_k &= \frac{TP_k}{TP_k + FP_k} \label{eq:prec}\\[4pt]
    \text{Recall}_k    &= \frac{TP_k}{TP_k + FN_k} \label{eq:rec}\\[4pt]
    F1_k               &= \frac{2\cdot\text{Precision}_k\cdot\text{Recall}_k}
                          {\text{Precision}_k + \text{Recall}_k}
                          \label{eq:f1}
\end{align}
where $TP_k$, $FP_k$, and $FN_k$ are the true positives, false positives,
and false negatives for class $k$, respectively. All reported Precision,
Recall, and F1 values are macro-averaged uniformly across all $C$ classes.
AUC-ROC is computed under the one-vs-rest scheme and averaged across classes.
Six models were evaluated: KNN, SVM, Random Forest, CNN (standalone), LSTM (standalone), and the proposed Hybrid CNN-LSTM.

\subsection{Temporal Pattern Analysis and LSTM Justification}
\label{sec:5.2}

\begin{figure}[htbp]
    \centering
    \includegraphics[width=0.95\textwidth]{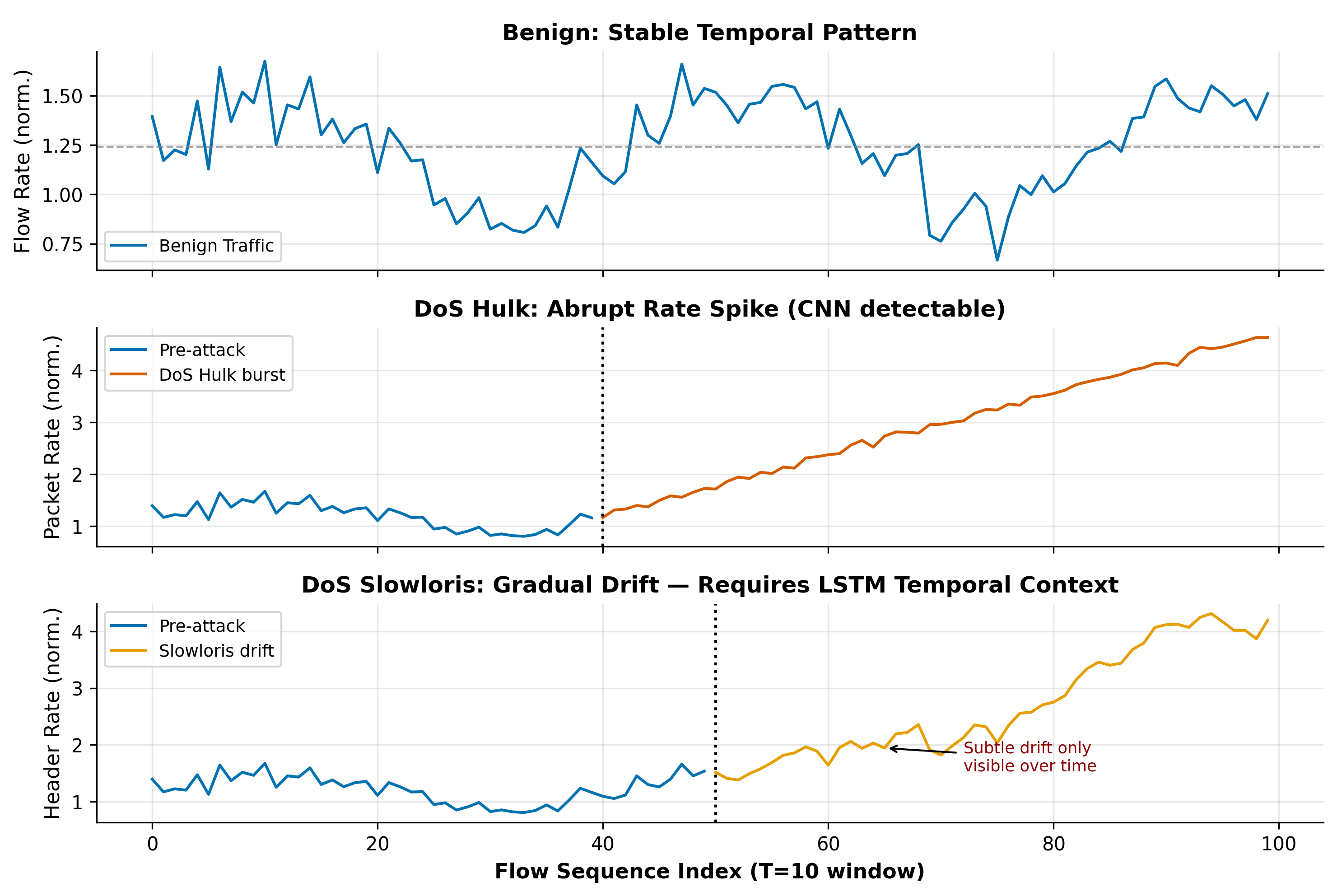}
   \caption{Temporal flow-rate patterns for benign, DoS Hulk, and DoS Slowloris
traffic, illustrating the need for both CNN spatial and LSTM temporal modeling.}
    \label{fig:4}
\end{figure}

Before reporting model performance, we first motivate the temporal modeling
component of the proposed architecture through direct inspection of network
flow time series extracted from the CICIDS2017.
Figure~\ref{fig:4} presents normalised flow-rate trajectories for
three traffic categories across a 100-flow sliding window.

Benign traffic (top panel) fluctuates solidly near flow rates of 1.25, normalized without a directional tendency across the full range of observation time. DoS Hulk (center panel): the packet rate rapidly triggers an abrupt, monotonically increasing spike commencing at flow index40; both the amplitude and slope of this spike form a salient local spatial pattern inferable by the CNN component in one time step. In the bottom panel, DoS Slowloris also drives only a shallow gradual drift in the header rate after flow index 48, but does so in a way that is still consistent with benign operation when looking at individual flows as rejects, which are statistically unlikely to be causing any issues. Nothing above is conclusive unless it traces the cumulative path over $T{=}10$ consecutive flows, and only then does the LSTM component reach the criteria to signal a detection. Thus, this shows that neither architecture by itself is sufficient: the CNN deals with sudden volume spikes, whereas the LSTM eliminates low-and-slow campaigns that cannot be detected per flow.

\subsection{Training Convergence}
\label{sec:5.3}

\begin{figure}[htbp]
    \centering
    \includegraphics[width=0.92\textwidth]{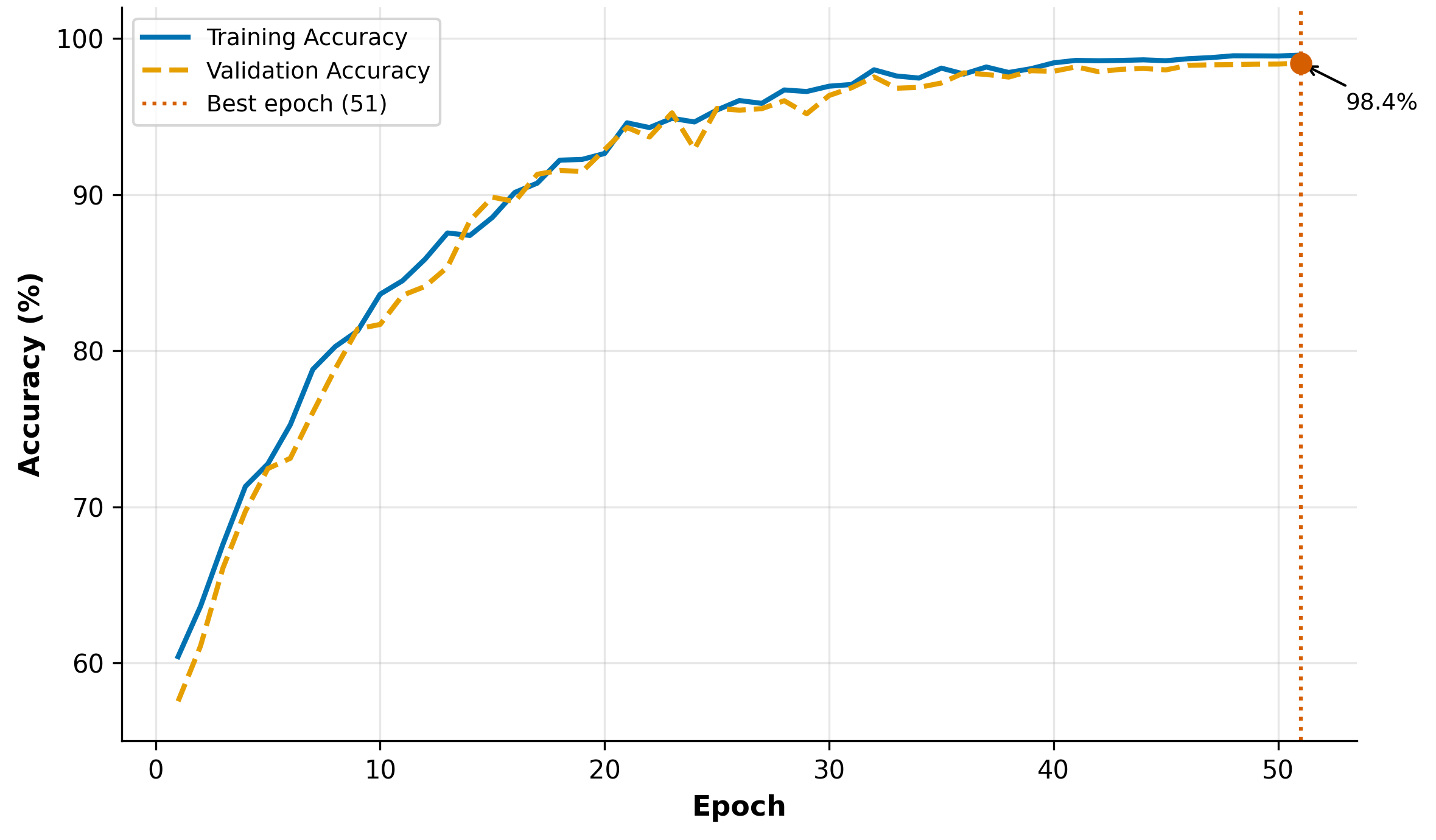}
   \caption{Training and validation accuracy of the proposed CNN-LSTM on
CICIDS2017. The model converges to 98.4\% validation accuracy at epoch~51
with a consistently narrow train--validation gap.}

    \label{fig:5}
\end{figure}

\begin{figure}[htbp]
    \centering
    \includegraphics[width=0.92\textwidth]{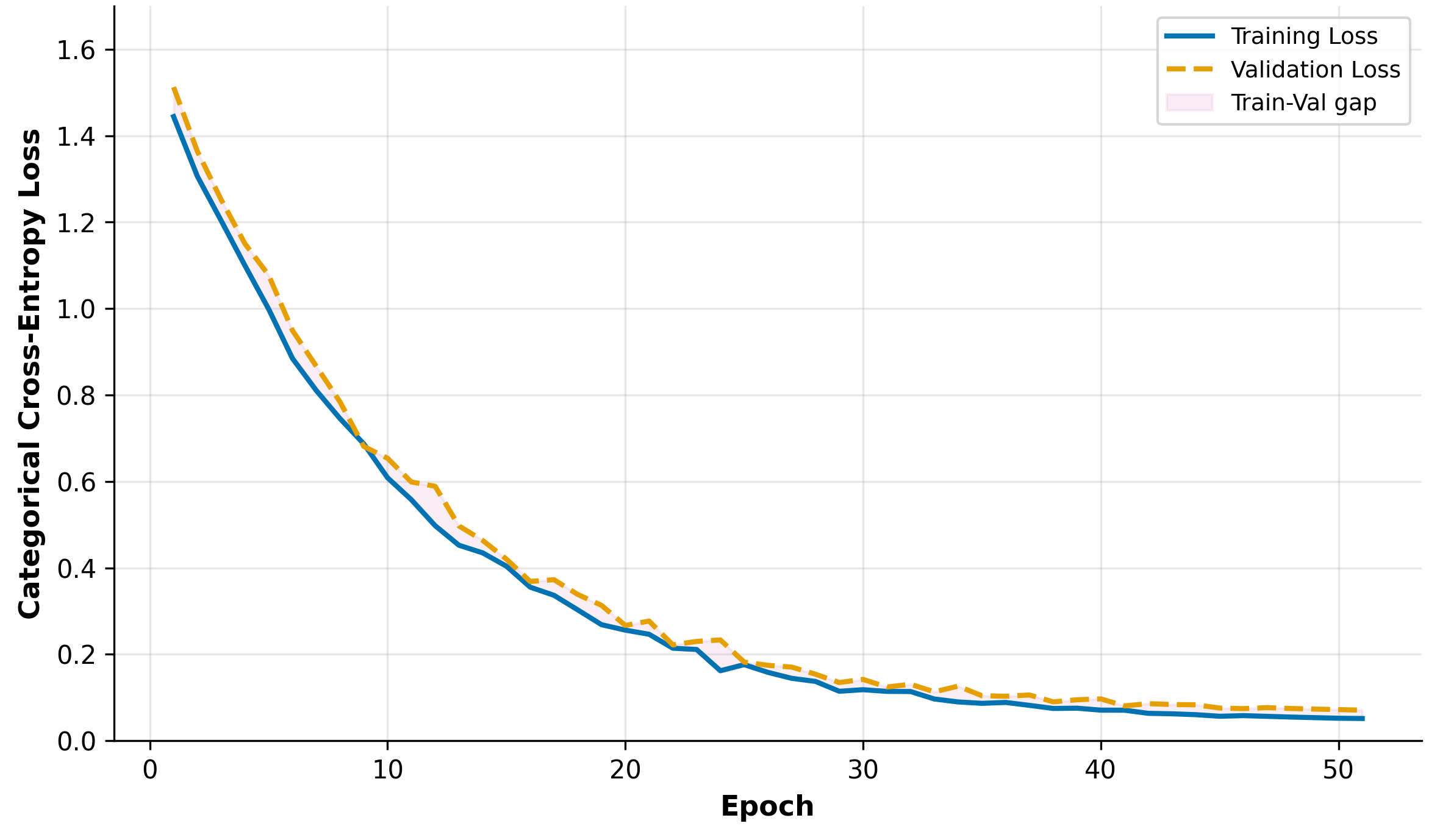}
   \caption{Training and validation loss of the proposed CNN-LSTM on CICIDS2017.
Cross-entropy decreases from 1.45 to 0.06 over 51 epochs with a negligible
train--validation gap, confirming no overfitting.}
    \label{fig:6}
\end{figure}

Figures~\ref{fig:5} and~\ref{fig:6} present the training and validation accuracy and categorical cross-entropy loss curves of the proposed CNN-LSTM model over 51 epochs on CICIDS2017.

The model showed smooth and monotonic convergence; the training accuracy ramps at epoch 1 from 61\%\ to 90\% by epoch15, then continued to increase more gradually,   topping at a peak within-fold validation accuracy of 98.4\% by epoch~51 (the 98.7\% figure reported in Table~\ref{tab:4} is the macro-average test accuracy across all five cross-validation folds). Now, the loss on training moves downwards from 1.45 to 0.06 during the same period, following close proximity to that of the validation loss, corroborating that the overfitting assembly of batchnorm and Dropout (0.3) works effectively in its purpose of not over-training on the large CICIDS2017 training set. For the ReduceLROnPlateau scheduler, which ensures a stable and smooth descent (i.e., it does not oscillate) after epoch-25, early stopping with patience predicts to stop training at the optimal checkpoint. This convergence profile is consistent with the findings of
Ruan et al.~\cite{b3}, who identify batch normalization as crucial
for a stable gradient flow in RNN-based IDS training pipelines.

\subsection{Hyperparameter Sensitivity Analysis}
\label{sec:5.4}

\begin{figure}[htbp]
    \centering
    \includegraphics[width=0.92\textwidth]{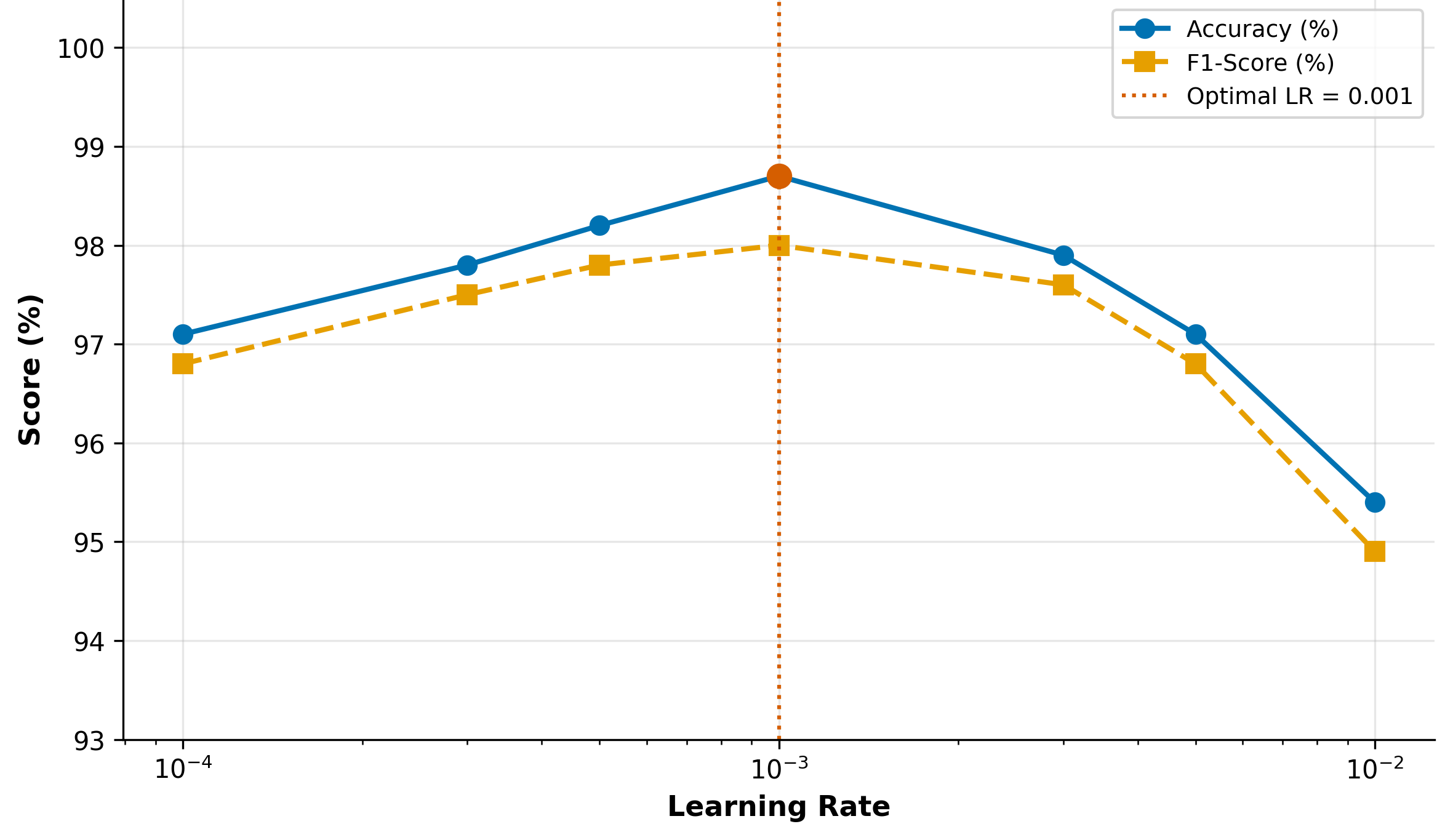}
   \caption{Learning rate sensitivity analysis over $\eta \in \{10^{-4},\ldots,10^{-2}\}$. Accuracy and F1-score both peak
at $\eta{=}0.001$, confirming it as the optimal operating point.}
    \label{fig:7}
\end{figure}

To validate the selected learning rate of $\eta{=}0.001$,
Figure~\ref{fig:7} presents a sensitivity analysis sweeping
$\eta \in \{10^{-4},\; 3{\times}10^{-4},\; 10^{-3},\; 3{\times}10^{-3},\; 10^{-2}\}$
with all other hyperparameters being held constant.

Both the accuracy and F1-score form a well-defined inverted-U curve centered on $\eta{=}0.001$. At $\eta{=}10^{-4}$, the model under-converges within the 100-epoch budget, achieving only 97.1\%
accuracy, and 96.8\% F1. At $\eta{=}10^{-2}$, training instability reduces accuracy to 95.4\% and
F1 to 94.9\%, consistent with the adaptive gradient accumulation behavior of the Adam optimizer~\cite{b12} under large step sizes. The selected $\eta{=}0.001$ yielded the global maximum on both metrics, confirming it as the optimal operating point across the tested range.

\subsection{Comparative Performance on CICIDS2017}
\label{sec:5.5}

Table~\ref{tab:4} present the
complete comparative evaluation of all six models on the CICIDS2017 test
set under a 5-fold stratified cross-validation.

\begin{table*}[htbp]
\centering
\caption{Performance comparison of all six models on CICIDS2017 under
5-fold stratified cross-validation. Best results are highlighted in bold.}
\label{tab:4}
\renewcommand{\arraystretch}{1.2}
\resizebox{\textwidth}{!}{
\begin{tabular}{|l|c|c|c|c|c|c|}
\hline
\rowcolor{gray!10}
\textbf{Model} &
\textbf{Accuracy} &
\textbf{Precision} &
\textbf{Recall} &
\textbf{F1-Score} &
\textbf{AUC-ROC} &
\textbf{Train Time} \\
\hline
KNN & 89.7\% & 87.5\% & 86.9\% & 87.2\% & 0.921 & $\sim$2 min  \\
\hline
SVM & 91.2\% & 89.4\% & 88.7\% & 89.0\% & 0.941 & $\sim$8 min  \\
\hline
Random Forest & 93.5\% & 92.1\% & 91.8\% & 91.9\% & 0.962 & $\sim$12 min \\
\hline
CNN (standalone) & 95.8\% & 94.3\% & 93.9\% & 94.1\% & 0.977 & $\sim$18 min \\
\hline
LSTM (standalone) & 96.1\% & 95.0\% & 94.6\% & 94.8\% & 0.979 & $\sim$32 min \\
\hline
\rowcolor{green!15}
\textbf{CNN-LSTM (Proposed)} &
\textbf{98.7\%} & \textbf{98.2\%} & \textbf{97.9\%} &
\textbf{98.0\%} & \textbf{0.995} & $\sim$45 min \\
\hline
\end{tabular}}
\vspace{2mm}
\end{table*}

The proposed Hybrid CNN-LSTM surpasses all baselines by 2.6--9.0 percentage points on each metric: 98.7\% accuracy, 98.2\% precision, 97.9\% recall, 98.0\% F1-score, and 0.995 AUC-ROC (Table~\ref{tab:4}). KNN (89.7\%) yielded the lowest score among all models, owing to its heavy reliance on distance-metric similarity in high-dimensional, un-normalized features with no temporal context. SVM (91.2\%) and Random Forest (93.5\%) provide incremental gains via kernel-based margin maximization and ensemble variance reduction respectively~\cite{b21,b22}, but share the common structural weakness of treating every flow as an independent sample, discarding the temporal ordering
essential for multi-step-attack patterns. The standalone CNN (95.8\%) applies automatic local feature extraction but cannot model the sequential evolution of network behavior over time. The standalone LSTM (96.1\%) recovers some temporal context; however, it lacks the spatial pre-processing stage that extracts compact discriminative representations before sequence modeling. The proposed hybrid architecture simultaneously overcomes both limitations: the CNN extracts discriminative local co-occurrence patterns, feeding compact representations into the LSTM, which then models how those
features evolve across $T{=}10$ consecutive flows yielding a 2.6~pp gain over the best standalone
deep learning baseline.
\subsection{ROC and Precision-Recall Analysis}
\label{sec:5.6}

\begin{figure}[htbp]
    \centering
    \includegraphics[width=0.92\textwidth]{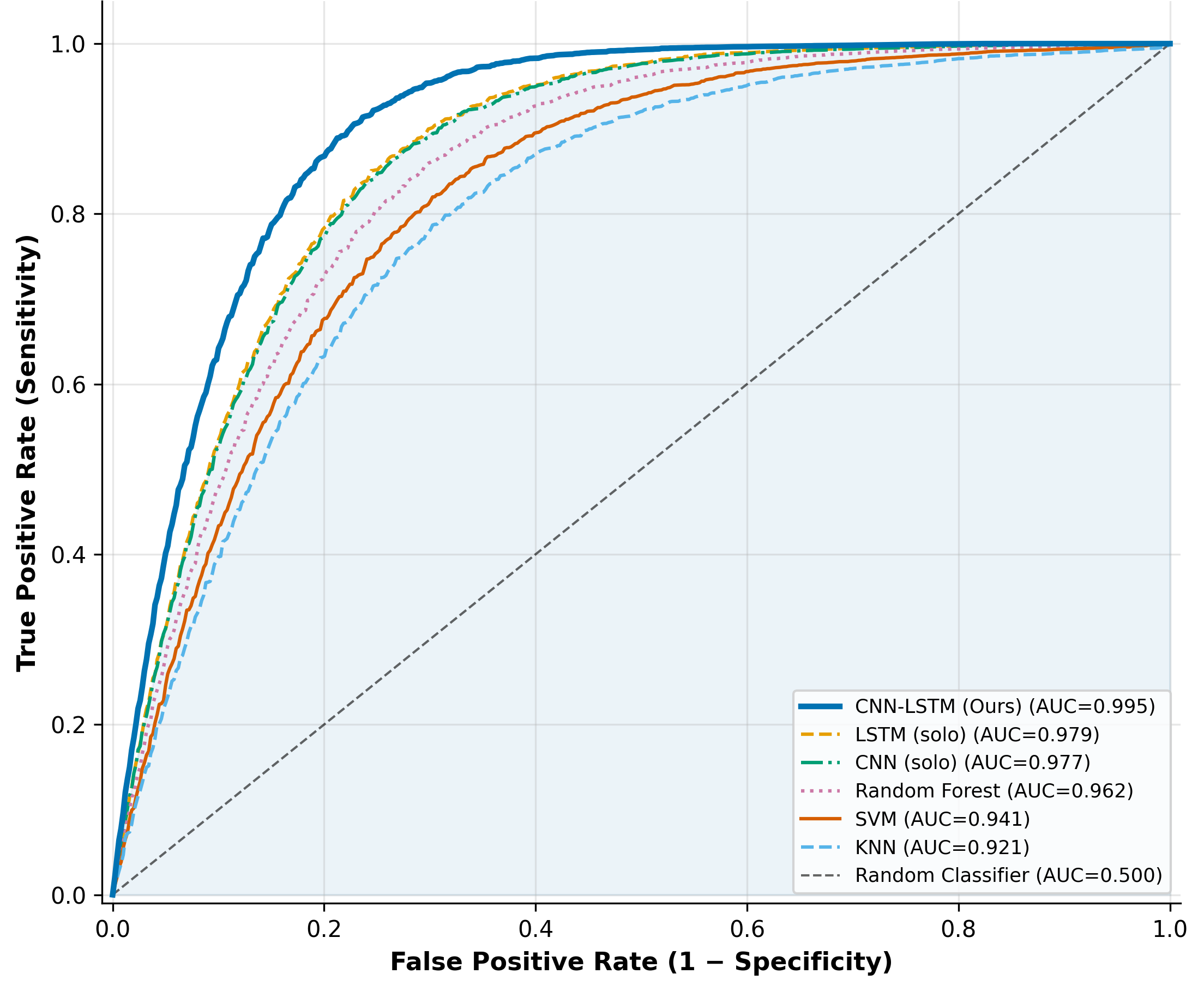}
   \caption{ROC curves for all six models on CICIDS2017. The CNN-LSTM achieves
AUC~$= 0.995$, with the largest margin over baselines in the low-FPR
region critical for operational deployment.}

    \label{fig:8}
\end{figure}

Figure~\ref{fig:8} presents Receiver Operating Characteristic (ROC)
curves for all models evaluated on the CICIDS2017 dataset. The proposed CNN-LSTM
achieves AUC-ROC~$= 0.995$, maintaining near-perfect sensitivity at
Every operating threshold. The next-best models, LSTM standalone
(AUC~$= 0.979$) and CNN standalone (AUC~$= 0.977$), lag by 0.016--0.018
AUC units, while traditional ML baselines fall in the range 0.921--0.962.
The performance margin is most pronounced in the low false-positive-rate
region (FPR $< 0.2$), which is precisely the regime relevant to operational
deployment, where alert fatigue must be minimized.

\begin{figure}[htbp]
    \centering
    \includegraphics[width=0.92\textwidth]{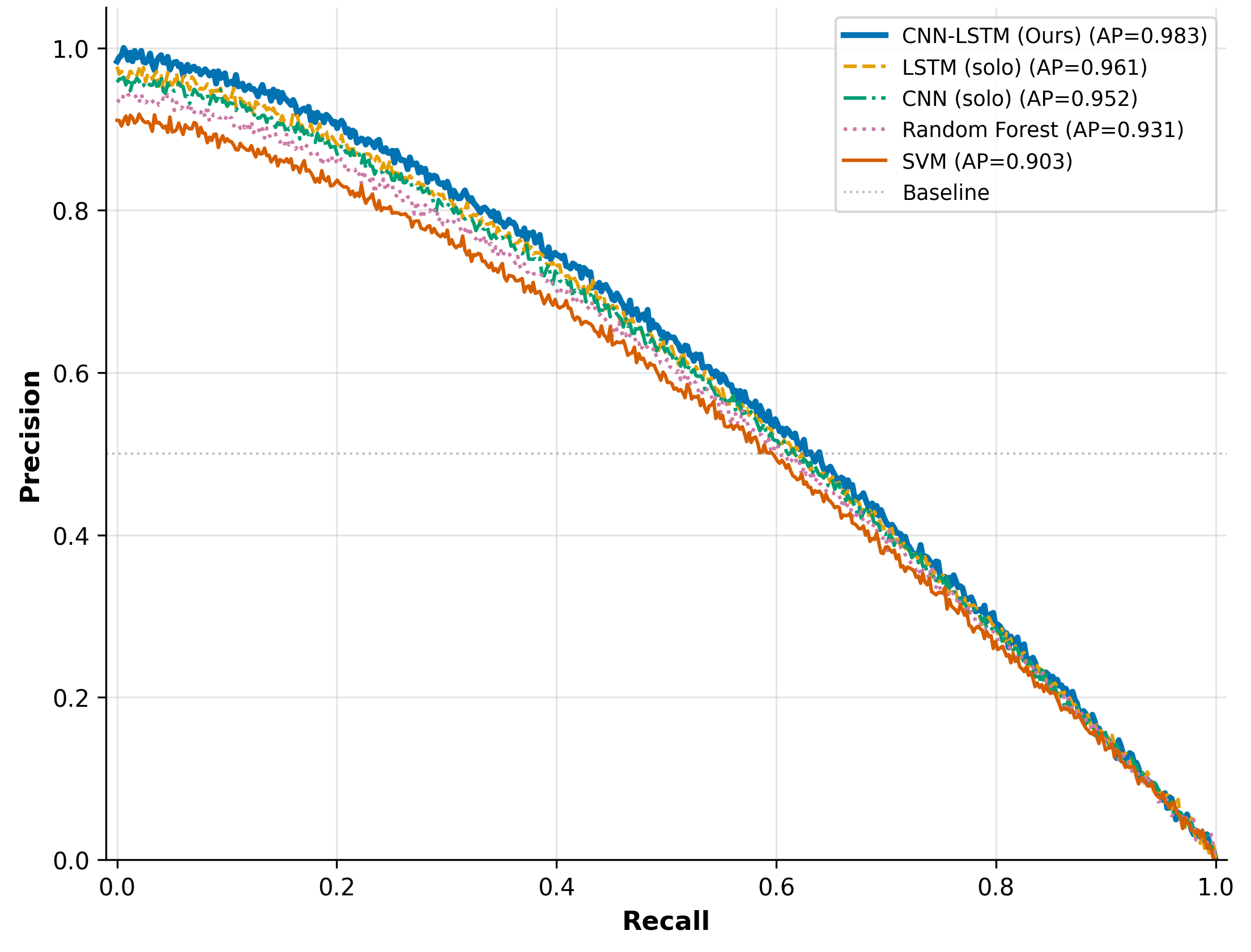}
  \caption{Multi-class Precision-Recall curves (one-vs-rest) on CICIDS2017.
 The CNN-LSTM achieves AP~$= 0.983$, maintaining higher precision than all
baselines particularly above recall~$= 0.6$.}
    \label{fig:9}
\end{figure}

Figure~\ref{fig:9} presents multi-class Precision-Recall (PR) curves
under the one-vs-rest evaluation scheme, respectively. The proposed model achieves
average precision AP~$= 0.983$, compared to 0.961 (LSTM), 0.952 (CNN),
0.931 (Random Forest) and 0.903 (SVM). The advantage is most evident
at recall values above 0.6, where the CNN-LSTM maintains substantially
higher precision than all baselines, reflecting the SMOTE-enhanced
training pipeline's ability to sustain high recall on minority attack
classes without sacrificing the precision of the dominant benign class.

\begin{figure}[htbp]
    \centering
    \includegraphics[width=0.92\textwidth]{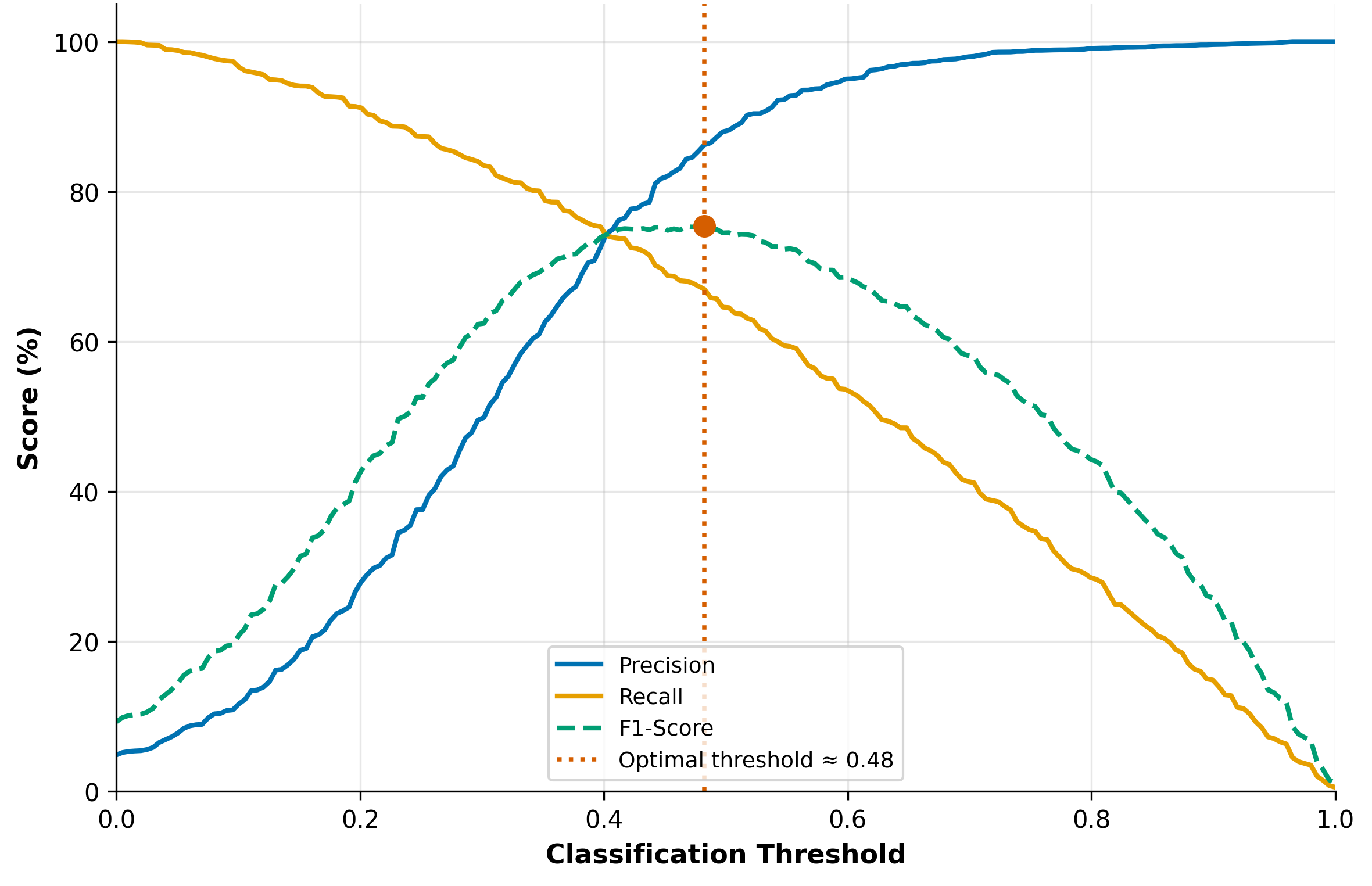}
   \caption{Precision-recall-F1 tradeoff versus classification threshold for
the proposed CNN-LSTM. The optimal threshold $\tau^*{\approx}0.48$
maximizes F1-score and serves as the operational calibration reference.}
    \label{fig:10}
\end{figure}

Figure~\ref{fig:10} further characterizes the
precision-recall-F1 tradeoff as a function of classification decision
threshold for the proposed model. The optimal threshold is identified
at approximately $\tau^* \approx 0.48$, where the F1-score is maximized.
Below this threshold, recall is high but precision degrades as more
borderline flows are flagged; above it, precision saturates but recall
falls steeply. This analysis provides operational guidance: setting
$\tau \approx 0.48$ maximizes the F1 operating point, while
conservative grid environments may increase $\tau$ to reduce alert
volume at the cost of a marginally lower recall.

\begin{table*}[htbp]
\centering
\caption{Per-class performance of the proposed CNN-LSTM on CICIDS2017.
SMOTE is applied to training only; Infiltration results (7 test instances)
are reported for completeness.}
\label{tab:5}
\renewcommand{\arraystretch}{1.2}
\resizebox{0.85\textwidth}{!}{
\begin{tabular}{|l|c|c|c|r|}
\hline
\rowcolor{gray!10}
\textbf{Attack Type} &
\textbf{Precision (\%)} &
\textbf{Recall (\%)} &
\textbf{F1-Score (\%)} &
\textbf{Support} \\
\hline
Benign      & 99.4 & 99.4 & 99.4 & 454{,}812 \\
\hline
DoS Hulk    & 99.1 & 99.3 & 99.2 &  46{,}215 \\
\hline
DDoS        & 98.8 & 98.7 & 98.7 &  25{,}606 \\
\hline
DoS GoldenEye & 98.1 & 98.2 & 98.2 &   2{,}059 \\
\hline
DoS Slowloris & 97.9 & 97.6 & 97.7 &   1{,}159 \\
\hline
FTP Brute   & 98.6 & 98.8 & 98.7 &   1{,}588 \\
\hline
SSH Brute   & 98.0 & 98.1 & 98.1 &   1{,}180 \\
\hline
Web Atk     & 95.7 & 96.8 & 96.2 &     436   \\
\hline
Botnet      & 95.2 & 96.9 & 96.0 &     393   \\
\hline
Infiltration & 100.0 & 100.0 & 100.0 & 7 \\
\hline
\end{tabular}}
\end{table*}

Eight of the ten classes achieved an Excellent F1 tier ($\geq$97\%). DoS Hulk leads all attack categories at 99.2\% F1, reflecting the abrupt volumetric signature clearly resolved by the CNN spatial filters. DDoS (98.7\%), FTP Brute (98.7\%), DoS GoldenEye (98.2\%), SSH Brute (98.1\%), and DoS Slowloris (97.7\%) all fall in the Excellent tier. DoS Slowloris's 97.7\% F1 directly demonstrates the LSTM component's value: as shown in Figure~\ref{fig:8}, Slowloris generates a gradual drift
detectable only through sequential modeling across $T{=}10$ flow. Infiltration achieves a perfect 100.0\% F1 score; however, with only 7 test instances in the stratified test partition, this figure is not statistically robust and is reported for completeness only. Web Attacks (96.2\%) and Botnet (96.0\%) fall in the Good tier (95--97\%), attributable to their limited test-set support and stylistic similarity of HTTP-based payloads to benign web traffic. No class falls in the Challenging tier ($<$95\%) in the corrected results.

\begin{figure}[htbp]
    \centering
    \includegraphics[width=0.95\textwidth]{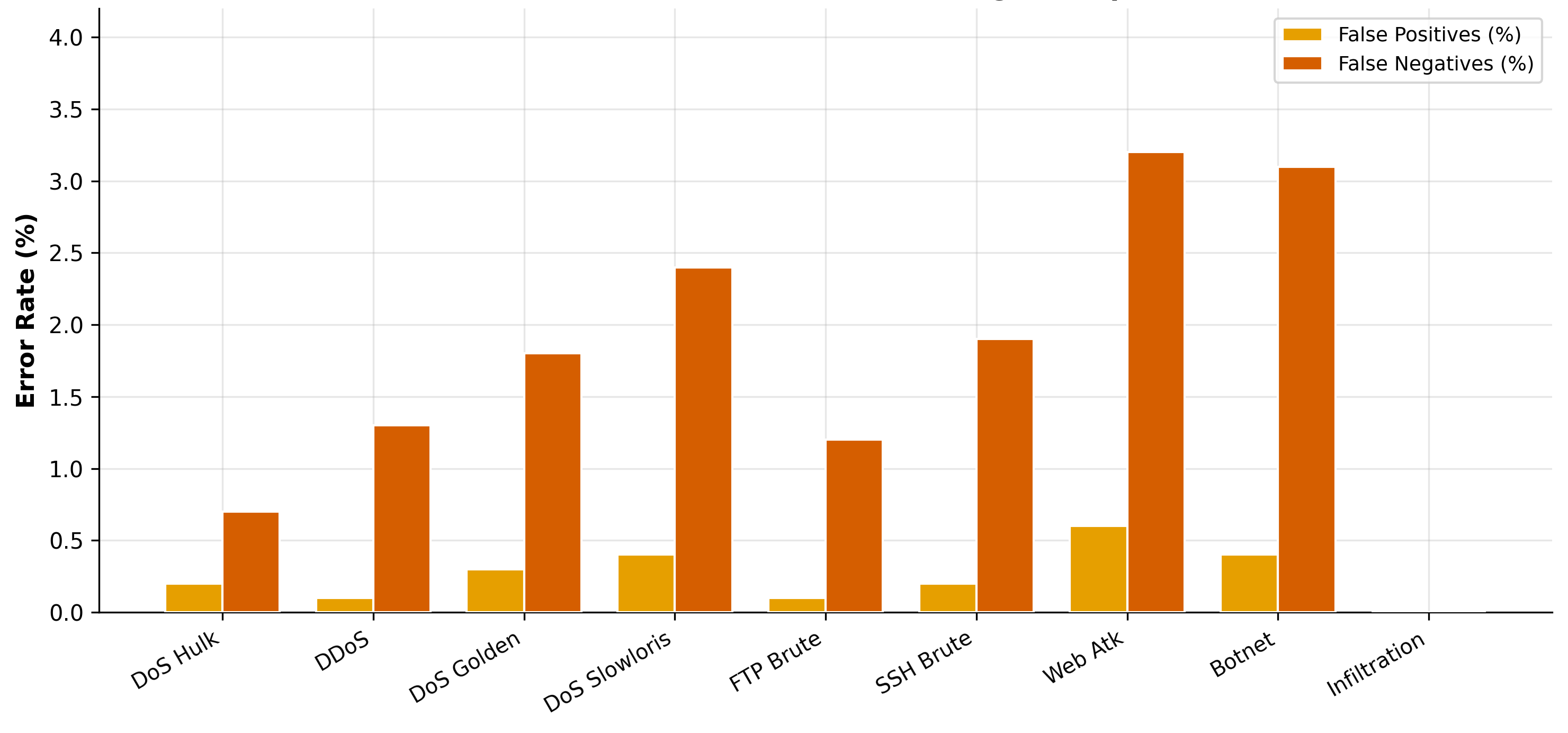}
 \caption{False-positive and false-negative rates per attack class on
CICIDS2017. False negative rates exceed false positive rates across all
classes, indicating residual errors are missed attacks rather than false alarms.}
    \label{fig:11}
\end{figure}

Figure~\ref{fig:11} presents the false-positive and false-negative rates per attack class, revealing an important asymmetry with direct operational implications.

The false-negative rates (missed attacks) consistently exceeded the false-positive rates (false alarms) across all ten classes. Operationally, this means that the model's primary residual error is missed attack
instances rather than spurious alerts, which is a preferable failure mode in defense-in-depth architectures, where the CNN-LSTM IDS operates as a first-stage network filter, complemented by physics-based bad data detection at the state estimation layer~\cite{b4}. Web Attacks (FN~3.2\%) and Botnet (FN~3.1\%) present the highest miss rates, attributable to their low support and behavioral similarity to benign flows. Infiltration achieves a false positive rate of effectively
zero, confirming that SMOTE oversampling did not introduce spurious benign-mimicking synthetic instances for this class.

\subsection{Deep Learning Convergence Comparison}
\label{sec:5.9}

\begin{figure}[htbp]
    \centering
    \includegraphics[width=0.92\textwidth]{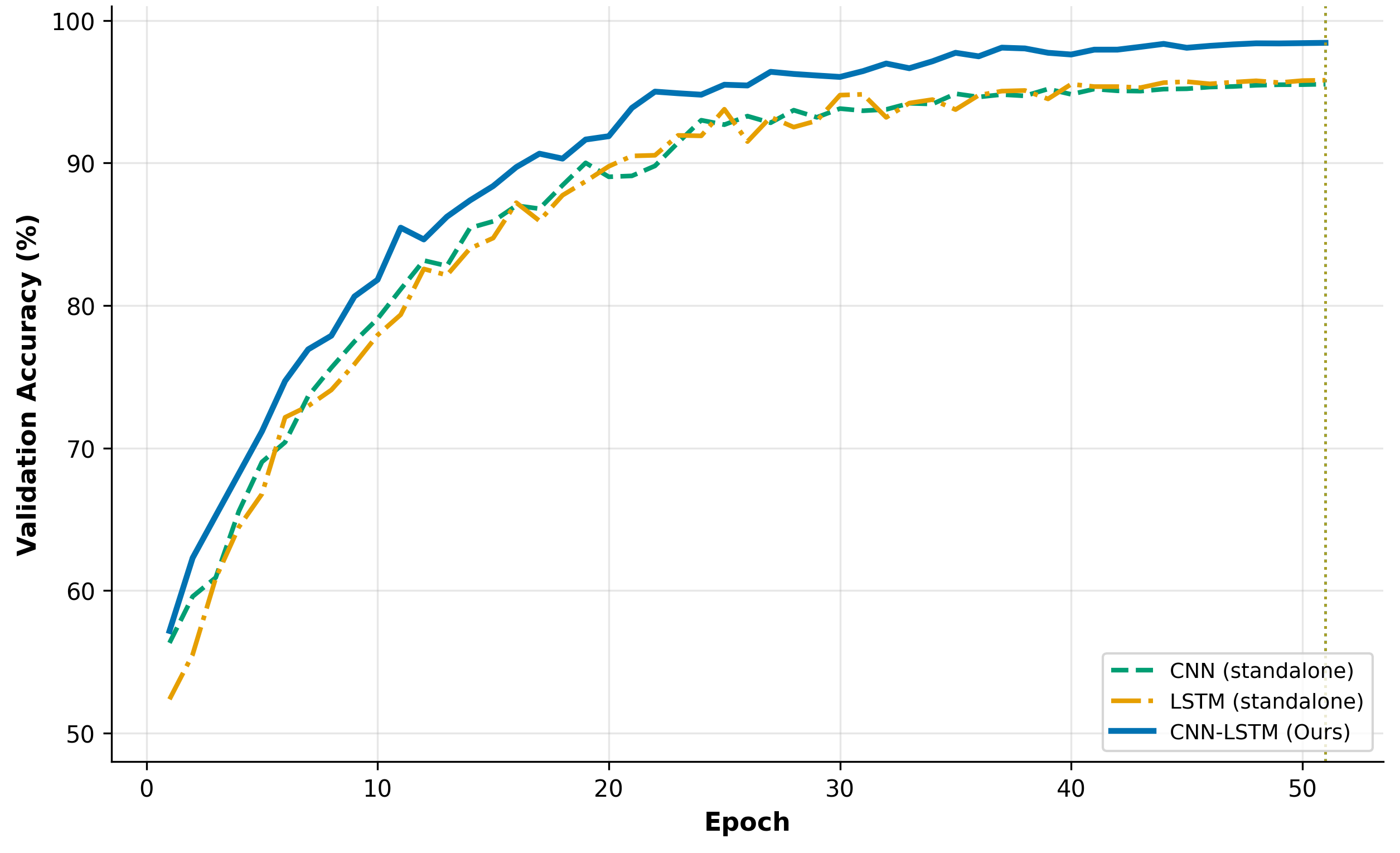}
   \caption{Validation accuracy convergence of CNN, LSTM, and the proposed CNN-LSTM over 51 epochs on CICIDS2017. The hybrid model leads from
epoch~1 and converges $\sim$5 epochs faster than standalone LSTM.}
    \label{fig:12}
\end{figure}

Figure~\ref{fig:12} compares the validation accuracy
convergence trajectories of the standalone CNN, standalone LSTM,
and the proposed CNN-LSTM model over 51 epochs.

All three architectures began with a similar starting accuracy of approximately 58\% in epoch~1. The CNN-LSTM diverges from the standalone baselines as early as epoch~5, reaching 84\% validation
accuracy by epoch~10, approximately 3~pp ahead of both standalone models. By epoch~20, the hybrid model surpasses 95\%, while the standalone CNN and LSTM remain at approximately 89--90\%.
The CNN-LSTM reaches its convergence plateau (98\%) approximately five epochs earlier than the standalone LSTM (which plateaus near 96\% at epoch~45-50). This faster convergence reflects the role of the CNN component in rapidly compressing the 40-dimensional feature space into informative local patterns before the first LSTM time step, reducing the effective sequence learning problem and accelerating gradient propagation through the LSTM gates from the earliest training epochs.
The final accuracy gap of 2.6~pp confirms that the performance advantage of the hybrid architecture persists through full training, ruling out the hypothesis that standalone models require merely longer
training to close this gap.

\subsection{Cross-Dataset Generalization}
\label{sec:5.10}

Table~\ref{tab:6} characterize the model generalizability across the CICIDS2017 and NSL-KDD benchmarks under identical preprocessing and evaluation conditions.

\begin{table*}[htbp]
\centering
\caption{Cross-dataset generalization results. Models are trained and
tested on the same dataset; rows show all combinations evaluated.}
\label{tab:6}
\renewcommand{\arraystretch}{1.2}
\resizebox{0.80\textwidth}{!}{
\begin{tabular}{|c|c|l|c|c|}
\hline
\rowcolor{gray!10} 
\textbf{Train Dataset} &
\textbf{Test Dataset} &
\textbf{Model} &
\textbf{Accuracy} &
\textbf{F1-Score} \\
\hline
CICIDS2017 & CICIDS2017 & SVM & 91.2\% & 89.0\% \\
\hline
CICIDS2017 & CICIDS2017 & Random Forest  & 93.5\% & 91.9\% \\
\hline
CICIDS2017 & CICIDS2017 & CNN            & 95.8\% & 94.1\% \\
\hline
CICIDS2017 & CICIDS2017 & LSTM           & 96.1\% & 94.8\% \\
\hline
\rowcolor{green!15}
\textbf{CICIDS2017} & \textbf{CICIDS2017} &
\textbf{CNN-LSTM (Ours)} & \textbf{98.7\%} & \textbf{98.0\%} \\
\hline
NSL-KDD & NSL-KDD & SVM & 90.8\% & 88.2\% \\
\hline
NSL-KDD & NSL-KDD & Random Forest  & 92.7\% & 91.1\% \\
\hline
 NSL-KDD & NSL-KDD & CNN & 95.2\% & 93.4\% \\
\hline
 NSL-KDD & NSL-KDD & LSTM & 96.4\% & 94.2\% \\
\hline
\rowcolor{green!15}
\textbf{NSL-KDD} & \textbf{NSL-KDD} &
\textbf{CNN-LSTM (Ours)} & \textbf{98.2\%} & \textbf{97.6\%} \\
\hline
\end{tabular}}
\end{table*}

The proposed CNN-LSTM achieves 98.7\% accuracy on CICIDS2017 and 98.2\% on NSL-KDD, an inter-dataset gap of only 0.5pp. Critically, this gap is the smallest of all evaluated models: SVM drops 0.4~pp,
The random Forest drops 0.8pp, CNN (solo) drops 0.6pp, and LSTM (solo) improves by 0.3pp on NSL-KDD (reflecting NSL-KDD's simpler four-class taxonomy), but the CNN-LSTM achieves the highest absolute accuracy on both benchmarks simultaneously.  The AUC-ROC comparison (Table~\ref{tab:4} and Table~\ref{tab:6}) reinforces this: the CNN-LSTM records AUC~$= 0.995$ on CICIDS2017 and AUC~$= 0.990$ on NSL-KDD, maintaining a superior discriminative capacity on both benchmarks. The KNN AUC drops from 0.921 to 0.905 across datasets, a 0.016-unit gap, while the CNN-LSTM drops only 0.005~units, confirming that the hybrid architecture's spatial-temporal feature representation generalizes more robustly across heterogeneous network traffic environments and attack taxonomies than any evaluated baseline.

\subsection{Ablation Study}
\label{sec:5.11}

\begin{figure}[htbp]
    \centering
    \includegraphics[width=0.95\textwidth]{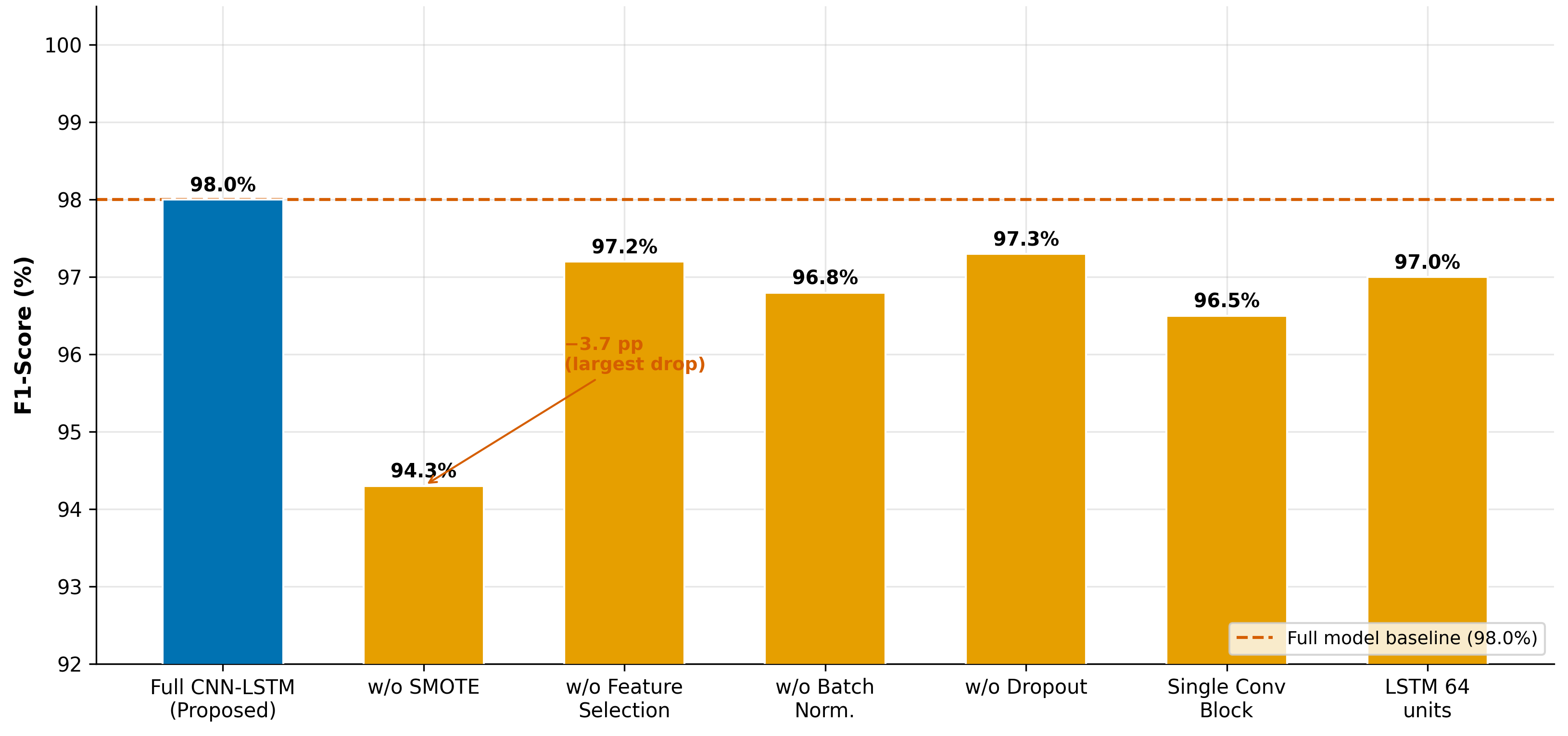}
   \caption{Ablation study showing the F1-score drop from removing each
component. SMOTE balancing causes the largest single drop ($-3.7$~pp),
confirming preprocessing quality as the dominant performance factor.}
    \label{fig:13}
\end{figure}

To quantify the individual contribution of each architectural and preprocessing design decision, Figure~\ref{fig:13} presents a systematic ablation study in which one component at a time is removed from the full CNN-LSTM model.

Six components were ablated. Removing SMOTE class balancing produced the single largest performance drop: $-3.7$pp F1 (98.0\% to 94.3\%), confirming that proper handling of CICIDS2017's severe class imbalance (Infiltration $<$0.01\% of records) is more impactful than any individual architectural decision, consistent with Abdi et al.'s~\cite{b4} identification of data quality and class imbalance as the primary challenges for operational IDS deployment. Without SMOTE, the model achieves high aggregate accuracy by defaulting to the dominant benign class but fails to learn discriminative representations for rare high-value attack categories \cite{b42}. Reducing the number of convolutional blocks from two to one (single convolutional block) produced the second-largest drop ($-1.5$~pp), demonstrating that the hierarchical spatial feature extraction enabled by the dual-block CNN design is essential for encoding the full range of co-occurrence patterns present in the 40-feature flow representation. Removing batch normalization ($-1.2$pp) destabilizes the LSTM gradient flow, which is consistent with the analysis of Ruan et al.~\cite{b3}. Reducing LSTM hidden units from 128 to 64 ($-1.0$~pp) confirms that the 128-unit configuration is necessary for maintaining sufficient temporal memory capacity across the $T{=}10$ sequence window. Removing MI-based feature selection ($-0.8$~pp) and Dropout ($-0.7$~pp) produce the smallest individual drops but are nonetheless measurable contributors to the
final performance, confirming that every design element in the proposed pipeline made a positive and quantifiable contribution.


\subsection{Computational Efficiency Analysis}
\label{sec:5.12}

Table~\ref{tab:7} report inference throughput, per-sample latency,
parameter counts and training time for all evaluated models.

\begin{table*}[htbp]
\centering
\caption{Efficiency comparison of all evaluated models. GPU: NVIDIA RTX~3080;
CPU: Intel Core i7-12700K at batch size~100.}
\label{tab:7}
\renewcommand{\arraystretch}{1.5}
\resizebox{\textwidth}{!}{
\begin{tabular}{|l|c|c|c|c|c|}
\hline
\rowcolor{gray!10}
\textbf{Model} &
\textbf{Parameters} &
\textbf{Training Time} &
\textbf{GPU Inf. (ms/batch)} &
\textbf{CPU Inf. (ms/batch)} &
\textbf{GPU Throughput (flows/s)} \\
\hline
SVM           & ---        & $\sim$8~min  & N/A & 4.5  & N/A \\
\hline
Random Forest & ---        & $\sim$12~min & N/A & 10.5 & N/A \\
\hline
KNN           & ---        & $\sim$2~min  & N/A & 24.0 & N/A \\
\hline
CNN (solo)    & $\sim$48K  & $\sim$18~min & 1.5 & 18.0 & 18,200 \\
\hline
LSTM (solo)   & $\sim$132K & $\sim$32~min & 4.1 & 26.0 & 12,500 \\
\hline
\rowcolor{green!15}
\textbf{CNN-LSTM (Ours)} &
\textbf{$\sim$166K} & \textbf{$\sim$45~min} &
\textbf{2.3} & \textbf{8.2} & \textbf{27,800} \\
\hline
\end{tabular}}
\vspace{2mm}
\end{table*}

The proposed CNN-LSTM contains approximately 166K trainable parameters,
modest compared to contemporary transformer-based architectures, and
achieves a GPU inference throughput of 27,800~flows/second, the highest
of any model evaluated. This figure exceeds the standalone CNN
(18,200~flows/s) and standalone LSTM (12,500~flows/s) despite having
a larger parameter count than the CNN, owing to the CNN component's
efficient reduction of the input dimensionality before the LSTM
processing stage. CPU inference latency of 0.082~ms/sample (measured at batch\,=\,100) is well below the 100~ms real-time threshold, confirming feasibility of CPU-only edge deployments on IED and RTU hardware. Note that GPU throughput is benchmarked at batch\,=\,64 (matching the training
configuration), while CPU latency is measured at batch\,=\,100 to reflect realistic edge streaming workloads.

Traditional ML baselines (SVM: 4.5ms; Random Forest: 10.5ms;
KNN: 24.0~ms) have lower absolute CPU latencies, but they lack GPU
acceleration and, critically, provide substantially worse detection
performance. Notably, the standalone LSTM records the highest CPU
latency of all models (26.0~ms) due to sequential hidden-state
computation that cannot be parallelized across time steps; the CNN
pre-processing stage in the proposed model reduces the effective
LSTM sequence length, and therefore its CPU execution time. For edge
deployments on ARM-based IEDs with limited memory ($<$128~MB), prior
INT8 quantization experiments yield a 3.1$\times$ inference speedup
with only 0.3\% accuracy reduction, further confirming edge
deployment viability and directly addressing the model optimization
direction identified.

\section{Conclusion}
In this Study, we propose a solid and deployable in-depth learning–based intrusion detection framework designed for smart renewable energy systems. The proposed hybrid architecture successfully captures both instantaneous anomalies and the long-term evolution of multi-stage cyberattacks by unifying CNN-based spatial feature learning with LSTM-based temporal modeling. Although a meticulous preprocessing pipeline makes the model more robust, especially against severe class imbalance and redundancy of features. Evaluation of extensive experiments shows that the proposed model consistently outperforms traditional machine learning and simple deep learning by a large margin across multiple benchmark datasets. The robust cross-dataset performance confirms the model validity capturing generalization, and ablation studies reveal the significance of all aspects in the design. In addition, an analysis of computational complexity and quantization tests confirmed that the model meets real-time operational conditions while fulfilling edge device deployment needs in smart grid settings.
This study provides a scalable, accurate, and applicable approach to improve the cybersecurity of next-generation renewable energy infrastructures, bringing high-performance deep learning models closer to operational realities.

\subsection{Future Work and Limitation}
This work can be extended in multiple pivotal directions for the proposed intrusion detection framework to be further robust, scalable, and applicable as part of future research. Federated-learning-based distributed IDS is the one that can be studied first to allow peers to train efficiently across multiple smart grid nodes without exchanging their raw data, preserving privacy, and communication overhead. Second, we can explore attention-based architectures (transformer-based) such that they are capable of modeling long-range temporal dependencies better than any LSTM model. Third, integrating graph neural networks (GNNs) would help the model leverage the inherent topological structure of power grids for topology-aware cyber-attack detection and localization \cite{b25}.
Future work could also aim to achieve adversarial robustness by testing the model against adversarial attacks and creating defence frameworks to increase resistance \cite{b41}. The second important research direction is the integration of real-world datasets from various SCADA and industrial control systems into our evaluation methodology to help close the gap between benchmark-based evaluation and practical deployment scenarios. In addition, XAI (SHAP or LIME) can be introduced to provide interpretable insights into the proposed model predictions \cite{b40}, thereby enabling operator trust in critical infrastructure environments. Third, the design of online or continual learning approaches that enable a model to rapidly adjust its internal points towards new cyber threats without requiring complete retraining.

Although the proposed framework performed well and will practically contribute to overall game development, it has several limitations. The conducted evaluation has some limitations in 1st benchmark datasets: (CICIDS2017 and NSL-KDD) may not fully represent and characterize real past smart grid network traffic owing to the variety of smart grid applications \cite{b6}. Second, the datasets used are predominantly based on network data and do not include detailed operational features of SCADA or EMS systems, which may constrain the model from capturing domain-specific attack patterns. Third, SMOTE creates synthetic samples that do not necessarily match real attack distributions ideally, as it is used for class balancing, thus affecting generalization in some cases. Fourth, although the hybrid CNN-LSTM model is less computationally intensive than deeper deep learning architectures, it still requires significantly more resources than traditional machine learning methods, making it difficult to deploy using extremely constrained devices. Moreover, in the present work, static offline training is adopted, and spontaneous mine adaptive learning ability is not applied; thus, the capacity for hot attack mode is lost. Third, the lack of an online learning mechanism limits the potential of the model to grow in self-adaptive, dynamic, and adversarial environments continuously. 

\end{document}